\documentclass[10pt,twocolumn,letterpaper]{article}

\usepackage[pagenumbers]{cvpr} 

\usepackage[T1]{fontenc}
\usepackage{float}

\definecolor{cvprblue}{rgb}{0.21,0.49,0.74}
\usepackage[pagebackref,breaklinks,colorlinks,allcolors=cvprblue]{hyperref}

\def\confName{CVPR}
\def\confYear{2027}

\title{LuxAvatar: Real-Time Relightable Head Avatars via Causal Neural Shading of 3D Intrinsic Proxies}

\author{{\normalsize Yipengjing Sun$^1$, Chenyang Wang$^1$, Boyao Zhou$^{\dagger,2}$, Jiapeng Zhu$^3$, Ka Leong Cheng$^3$, Shengping Zhang$^{*,1}$}\\
{\normalsize $^1$Harbin Institute of Technology\quad $^2$Tsinghua University\quad
$^3$HKUST} \\
}
\hypersetup{pdfauthor={Yipengjing Sun, Chenyang Wang, Boyao Zhou, Jiapeng Zhu, Ka Leong Cheng, Shengping Zhang}}

\begin{document}

\twocolumn[{
  \begin{@twocolumnfalse}
    \maketitle
    \begin{figure}[H]
      \captionsetup{skip=10pt}
      \hsize=\textwidth
      \centering
      \vspace{-10mm}
      \includegraphics[width=0.85\textwidth]{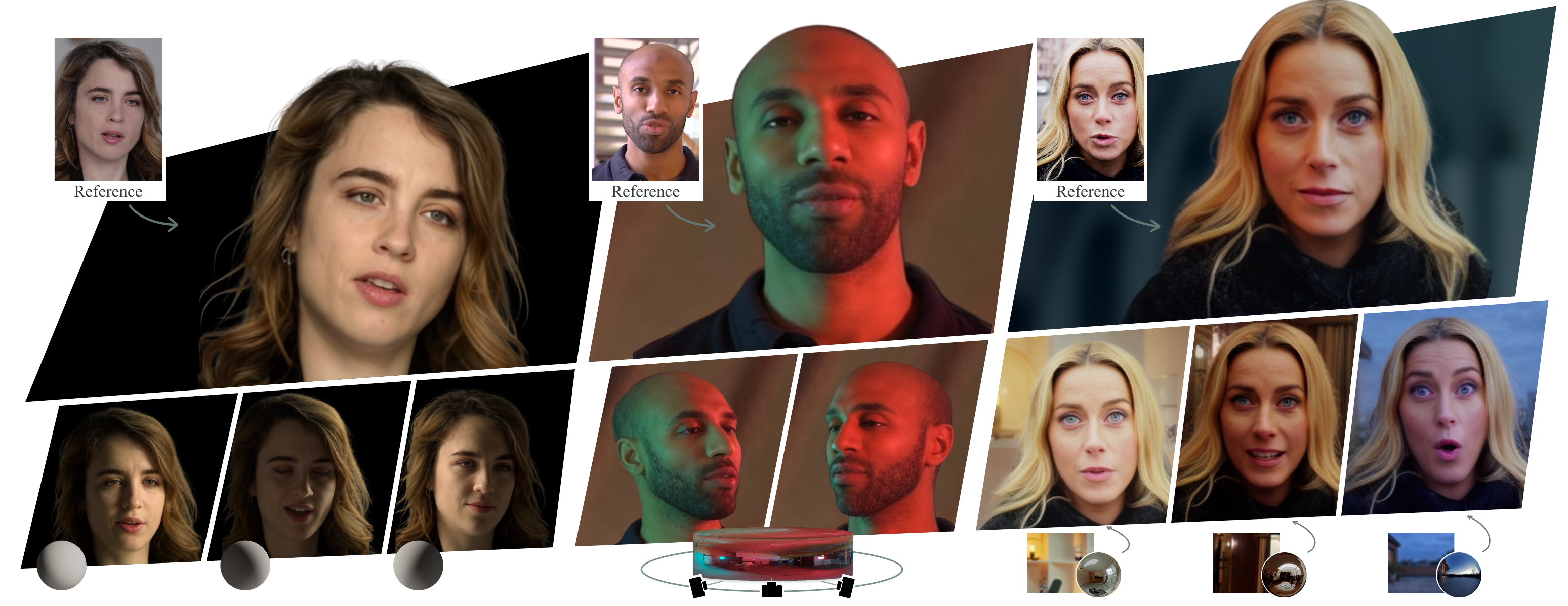}
      \vspace{-2mm}
      \caption{\textbf{Overview of LuxAvatar.}
From a single portrait, our method enables real-time head animation under dynamic illumination, scene-consistent background harmonization, and 3D viewpoint control.}
      \label{fig:teaser}
    \end{figure}
  \end{@twocolumnfalse}
}]

\begingroup
  \renewcommand{\thefootnote}{\fnsymbol{footnote}}
  \footnotetext[2]{Project lead.}
  \footnotetext[1]{Corresponding author.}
\endgroup

\begin{abstract}
We present LuxAvatar, a feed-forward framework for real-time head avatar animation and relighting from a single portrait.
To achieve both 3D controllability and complex facial light transport, we factorize avatar synthesis into an animatable 3D intrinsic Gaussian proxy and a causal autoregressive neural shader.
This proxy disentangles canonical geometry and reflectance attributes, rapidly rasterizing deformed Gaussians into view- and motion-consistent PBR buffers.
Conditioned on these buffers, our neural shader formulates forward rendering as a conditional video diffusion process, bypassing explicit PBR computation to synthesize realistic shading effects under target illumination.
We achieve real-time streaming through heterogeneous distillation from a pretrained multi-step bidirectional relighting teacher to a two-step causal autoregressive neural shader, inheriting broad synthetic lighting priors while minimizing conditioning and decoding latency.
To further specialize these priors to human heads, we construct a hybrid relighting dataset for domain-specific fine-tuning, capturing authentic facial dynamics and physically grounded light transport.
Extensive experiments show that our framework enables 3D-controllable, real-time portrait animation and relighting, outperforming state-of-the-art methods in visual fidelity, motion control, and lighting accuracy.
\end{abstract}
\section{Introduction}
\label{sec:intro}
Creating animatable and relightable head avatars is crucial for telepresence, immersive communication, and digital content creation.
Seamless integration into virtual environments demands not only faithful facial motion, but also photorealistic appearance that adapts naturally to novel illumination.
However, convincing head relighting is inherently difficult, as it requires modeling the intricate light transport among dynamic facial geometry, complex skin reflectance, and environmental lighting.
Without costly acquisition setups or heavy rendering overhead, capturing and representing these complex effects with high visual fidelity remains a fundamental challenge.

Traditional approaches~\cite{debevec2000acquiring, ma2007rapid, ghosh2011multiview} to relightable head modeling either reconstruct facial meshes and reflectance maps for physically based rendering or linearly combine images captured under basis lighting. 
However, they often struggle with complex structures such as hair and detailed appearance.
Subsequent neural volumetric head models~\cite{yang2023towards,sarkar2023litnerf,jiang2023nerffacelighting,xu2024artist,mei2024holo,rao2024lite2relight,rao20253dpr} capture fine geometry and lighting effects via learned volumetric fields, at the cost of slow ray marching.
Recently, relightable 3D Gaussian Splatting (3DGS) methods~\cite{saito2024relightable,li2024uravatar,becominglit,lv2025gsheadrelight,xu2026relightanyone} have accelerated rendering by rasterizing explicit Gaussian primitives augmented with material properties or learned light transport.
Despite these rendering advances, reconstructing a new subject still requires time-consuming per-subject optimization and faces a fundamental trade-off in data acquisition.
Light stages provide controlled illumination but entail prohibitive hardware costs, whereas monocular approaches~\cite{wang2023sunstage, 2023flare, zhang2025hravatar, xu2025monocular} operate under unknown lighting via ill-posed inverse rendering,  often suffering from baked-in shading residuals and simplified analytical shaders that struggle to model realistic light transport.

To enable instant avatar creation from a single portrait, recent feed-forward Gaussian head models~\cite{chu2024generalizable,he2025lam,zhao2026one,peng2026flexavatar,kirschstein2026flexavatar,wu2026uika,youwang2026fica} learn cross-identity priors for rapid 3D reconstruction and animation.
However, these methods focus solely on motion control under baked-in ambient lighting, leaving the avatars non-relightable.
Alternatively, 2D conditional diffusion models~\cite{liang2024caphuman, jang2025controlface} directly synthesize portraits under target motion and lighting, yet struggle with precise 3D control and cross-frame identity consistency~\cite{guo2025high}.
Moreover, their multi-step denoising incurs high computational overhead, prohibiting real-time interaction.
%
%

To address these challenges, we introduce LuxAvatar, a feed-forward framework for real-time streaming head animation and relighting from a single image, as shown in Fig.~\ref{fig:teaser}.
Our key insight is to couple an animatable 3D intrinsic Gaussian proxy with a causal autoregressive neural shader that models complex light transport from 4D-consistent PBR buffers in real time.
Specifically, given a single reference image, we first reconstruct 3D Gaussians anchored on a canonical UV manifold in a feed-forward manner, augmenting each primitive with decoupled geometric and material attributes, including surface normals, base color, and roughness.
Driven by target FLAME and camera parameters, these canonical primitives are deformed and rapidly rasterized into view- and motion-consistent PBR buffers.
Such a Gaussian proxy imposes a prior on 3D consistency and temporal material coherence.

Crucially, since the proxy already accounts for 3D geometry and viewpoint changes, our neural shader primarily addresses how to model intricate light transport efficiently and accurately.
Rather than explicitly formulating analytical PBR equations, we cast forward shading as a conditional video diffusion process that directly synthesizes complex illumination effects, such as subsurface scattering, specular highlights, and soft shadows. 
However, standard video diffusion suffers from multi-step sampling overhead and bidirectional frame dependencies. 
To achieve both high fidelity and efficiency, we distill a pre-trained multi-step bidirectional teacher tailored for video relighting into a two-step causal autoregressive generator with lightweight channel-wise conditioning and fast latent decoding.
%
%
Since the pre-trained teacher is designed for general relighting, we subsequently fine-tune the distilled generator on portrait relighting data to specialize it for human portraits.

Both training our feed-forward Gaussian proxy and fine-tuning the distilled shader require paired portrait videos under varying illumination.
To overcome the high cost of acquiring such paired data at scale, we further construct a hybrid dataset  from three complementary sources:  
%
%
%
(i) synthetic intrinsic supervision for the proxy and relit video targets for the shader, (ii) dynamic portraits under static lighting via reverse reference editing to provide real video targets with authentic facial dynamics, and (iii) static portraits under rotating illumination from public OLAT captures for physically grounded light transport.
%

Our main contributions can be summarized as follows:
\begin{itemize}
    \item We present LuxAvatar, a feed-forward framework for real-time streaming head avatar animation and relighting from a single portrait, supporting explicit motion and viewpoint control alongside photorealistic light transport without per-subject optimization.
    \item We introduce an animatable intrinsic Gaussian proxy that predicts decoupled canonical geometry and material attributes, providing view- and motion-consistent PBR buffers as explicit 4D guidance for neural shading.
    \item We develop a two-step causal neural shader via heterogeneous distillation and curate a hybrid dataset that adapts the general relighting prior with real portrait dynamics and OLAT supervision, enabling efficient streaming with realistic facial appearance and dynamic shading.
\end{itemize}

\section{Related Work}
\label{sec:related_work}

\subsection{Feed-Forward 3D Head Avatars}
Feed-forward head avatar methods learn cross-identity priors to reconstruct animatable heads from one or a few images without per-identity optimization.
Early radiance-field head avatars~\cite{li2023one, li2023generalizable, chu2024gpavatar, ye2024real3dportrait, deng2024portrait4d, deng2024portrait4dv2, tran2024vodo3d, tran2024vodoxp} represent identity and motion through deformable fields or triplanes, supporting novel-view animation at the cost of expensive ray sampling for volumetric rendering.

Recently, 3D Gaussian representations have mitigated this rendering bottleneck through efficient splatting, while exploring diverse animation and rendering designs.
To model facial dynamics, expression-conditioned Gaussian predictors~\cite{kirschstein2025avat3r, ji2026fastgha, peng2026flexavatar, kirschstein2026flexavatar, youwang2026fica} directly regress motion-dependent Gaussian attributes.
Alternatively, canonical Gaussian models~\cite{he2025lam, sun2026fhavatar} achieve efficient animation via FLAME-based rigging and direct RGB splatting.
Along this line, UIKA~\cite{wu2026uika} constructs a Gaussian avatar from unposed images by leveraging facial correspondences and attention across screen and UV spaces for high-quality animation.
Beyond direct color splatting, GAGAvatar~\cite{chu2024generalizable} enhances image synthesis by coupling explicit Gaussian feature splatting with a neural image decoder.
Despite their efficiency, these methods bake illumination into reconstructed appearance without explicit reflectance--illumination disentanglement, thereby precluding~relighting.
%
\subsection{Relightable 3D Head Models}
Methods using light stages or other calibrated capture rigs~\cite{bi2021deep, xu2024artist, yang2023towards, sarkar2023litnerf, saito2024relightable, becominglit} build detailed relightable head models from controlled multi-view observations, but require dedicated capture and per-subject training.
To relax capture requirements, inverse-rendering approaches~\cite{wang2023sunstage, xu2025monocular, 2023flare, zhang2025hravatar} recover geometry, reflectance, and lighting from monocular observations through subject-specific optimization.
Given only a single image, parametric face reconstruction methods~\cite{yamaguchi2018high, smith2020morphable, lattas2020avatarme, lattas2021avatarme++, dib2021towards, caselles2023sira, papantoniou2023relightify, dib2024mosar} leverage morphable priors to infer these intrinsic properties, yet the decomposition remains ambiguous under unknown illumination.

Beyond surface reflectance recovery, generative and feed-forward head models~\cite{tan2022volux, jiang2023nerffacelighting, ranjan2023facelit, deng2024lumigan, mei2024holo, lv2025gsheadrelight, nelf, rao2024lite2relight, rao20253dpr} learn cross-identity priors over volumetric or Gaussian representations to support joint view synthesis and relighting.
However, to achieve high-fidelity facial animation, existing frameworks still rely on costly test-time optimization for each new subject.
VRMM~\cite{yang2024vrmm} fits an animatable volumetric head model to a single image via latent inversion,
URAvatar~\cite{li2024uravatar} adapts a universal Gaussian avatar to a phone video with joint lighting estimation,
and RelightAnyone~\cite{xu2026relightanyone} reconstructs a neutral-expression head from a single image through identity and lighting inversion.
%

\subsection{Generative Relighting and Rendering}
Generative portrait relighting methods~\cite{ponglertnapakorn2023difareli, ponglertnapakorn2025difareli++, chaturvedi2025synthlight, georelight, chen2026polar} synthesize novel illumination while preserving the observed pose and expression.
Similarly, portrait harmonization approaches~\cite{pandey2021total, ren2024relightful, zhang2025scaling} adapt foreground illumination to match replacement backgrounds.
%
Portrait video relighting methods~\cite{he2024diffrelight, mei2025lux, zhang2026pixelcube} extend image relighting to dynamic facial performances using flat-lit sequences that retain the recorded motion.
They promote temporal coherence through dynamic Gaussian reconstruction and temporal blending, or by jointly denoising multiple frames with video diffusion.
More broadly, diffusion-based neural renderers~\cite{zeng2024rgb, chen2025uni, fang2026v, liang2025diffusionrenderer, relitlive, ma2026livelight} synthesize photorealistic RGB images or videos under target illumination, using conditions such as geometry and material buffers, depth-aware irradiance maps, and RGB references.

For joint control of pose, expression, and lighting, 3DMM-guided diffusion models~\cite{ding2023diffusionrig, liang2024caphuman, jang2025controlface} use parametric face renderings as spatial conditions for portrait synthesis.
LCVD~\cite{guo2025high} extends this paradigm to video diffusion, conditioning on reference appearance and FLAME shading sequences to synthesize relightable portrait animation over overlapping temporal windows.
However, many existing video relighting and rendering frameworks still incur substantial latency from multi-step denoising.

\begin{figure*}[t]
    \centering
    \includegraphics[width=\linewidth]{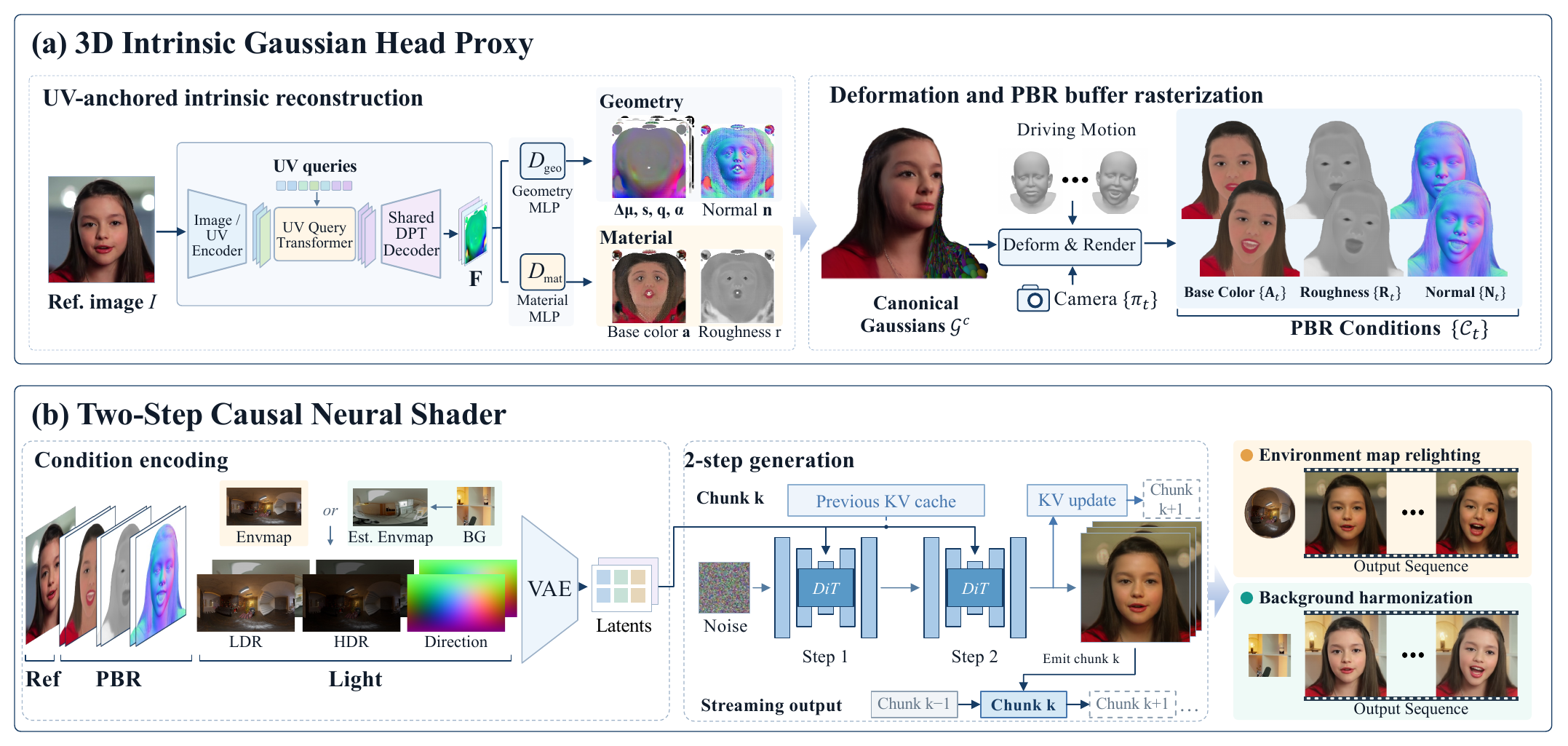}
    \caption{\textbf{Overview of LuxAvatar.} 
    (a) \textbf{Intrinsic Gaussian Head Proxy:} Given a single reference image $I$, our feed-forward network predicts decoupled geometry and material attributes on a canonical UV manifold. Driven by target FLAME and camera parameters, the canonical Gaussians are deformed and rasterized into 4D-consistent PBR buffers $\{\mathcal{C}_t\}$. 
    (b) \textbf{Causal Neural Shader:} Conditioned on $\{\mathcal{C}_t\}$, the reference image, and target scene illumination, our distilled causal neural shader autoregressively streams photorealistic relit video chunks in just two denoising steps, while reusing KV caches across chunks to maintain temporal coherence.
    }
    \label{fig:method_overview}
\end{figure*}
\section{Method}
\label{sec:method}
Given a reference image $I$, a sequence of driving FLAME parameters, and target camera parameters, our goal is to synthesize photorealistic, 3D-consistent portrait videos conditioned on either a target environment map for relighting or a target background for harmonization.
As illustrated in Fig.~\ref{fig:method_overview}, LuxAvatar consists of two core components: 
(1) a feed-forward intrinsic Gaussian head proxy (Sec.~\ref{sec:head_representation}) that lifts the input portrait into an animatable 3D representation to provide view- and motion-consistent PBR buffers, and 
(2) a causal autoregressive neural shader (Sec.~\ref{sec:video_renderer}) first distilled from a pre-trained bidirectional teacher for two-step streaming and relighting priors and then adapted on paired portrait data to achieve realistic facial shading and enable background harmonization.
To effectively supervise the intrinsic proxy and support paired portrait adaptation without costly dynamic light stage captures, we further detail our hybrid training dataset construction in Sec.~\ref{sec:data_construction}.

\subsection{3D Intrinsic Gaussian Head Proxy}
\label{sec:head_representation}
To provide explicit 3D geometric and material guidance from a single reference image $I$, we reconstruct a canonical head proxy parameterized by \(N\) 3D Gaussians~\cite{kerbl20233d}, where each primitive is augmented with intrinsic PBR attributes:
\begin{equation}
    \mathcal{G}^{c}
    = \left\{(\boldsymbol{\mu}_i,\mathbf{s}_i,\mathbf{q}_i,\alpha_i,
    \mathbf{n}_i,\mathbf{a}_i,r_i)\right\}_{i=1}^{N}.
    \label{eq:canonical_gaussians}
\end{equation}
Here, $\boldsymbol{\mu}_i$, $\mathbf{s}_i$, $\mathbf{q}_i$, and $\alpha_i$ denote the canonical 3D center, scale, orientation quaternion, and opacity of the $i$-th Gaussian, while $\mathbf{n}_i \in [-1,1]^3$, $\mathbf{a}_i \in [0,1]^3$, and $r_i \in [0,1]$ represent its unit normal, base color, and roughness.

\noindent\textbf{UV-anchored intrinsic reconstruction.}
We organize the canonical Gaussians on a 2D UV manifold anchored to an identity-shaped FLAME template~\cite{FLAME:SiggraphAsia2017}.
Given the reference image $I$, we first employ a frozen pretrained dense correspondence predictor to map foreground pixels onto the canonical UV grid, averaging colors within each occupied texel and zero-padding unobserved regions to form a partial UV texture $I_{\mathrm{uv}}$.
Both $I$ and $I_{\mathrm{uv}}$ are encoded by a shared frozen DINOv3~\cite{simeoni2025dinov3} backbone, followed by separate trainable adapters that fuse multi-layer intermediate features into image-space and UV-space token streams.
Learnable queries anchored at canonical UV coordinates then integrate both token streams via a transformer, after which a DPT-style decoder~\cite{ranftl2021vision} combines intermediate query features with RGB cues from $I_{\mathrm{uv}}$ to produce a dense canonical feature map $\mathbf{F}$.
Finally, for the $i$-th Gaussian at UV coordinate $\mathbf{u}_i$, two decoupled prediction heads regress its geometric and material attributes in parallel:
\begin{equation}
    \begin{aligned}
        D_{\mathrm{geo}}(\mathbf{F}(\mathbf{u}_i))
        &= (\Delta\boldsymbol{\mu}_i,\mathbf{s}_i,\mathbf{q}_i,\alpha_i,\mathbf{n}_i),\\
        D_{\mathrm{mat}}(\mathbf{F}(\mathbf{u}_i))
        &= (\mathbf{a}_i,r_i).
    \end{aligned}
    \label{eq:canonical_attribute_maps}
\end{equation}
Each canonical center is defined as $\boldsymbol{\mu}_i=\mathbf{v}_i+\Delta\boldsymbol{\mu}_i$, where $\mathbf{v}_i$ is the 3D surface point barycentrically interpolated from the FLAME mesh at $\mathbf{u}_i$.
%

\noindent\textbf{Deformation and PBR buffer rasterization.}
Given the driving FLAME motion parameters at frame $t$, we animate the canonical proxy via expression blendshapes and linear blend skinning (LBS), utilizing deformation bases and skinning weights barycentrically interpolated at $\mathbf{u}_i$.
Specifically, this deformation process updates the canonical centers $\boldsymbol{\mu}_i$ and transforms the normals $\mathbf{n}_i$ into the posed space using blended linear transforms followed by unit-length normalization, whereas the intrinsic material attributes ($\mathbf{a}_i, r_i$) remain unchanged. This yields the animated proxy $\mathcal{G}_t$ with view- and motion-consistent reflectance.
Under the target camera $\pi_t$, we then rasterize $\mathcal{G}_t$ into per-frame PBR buffers via fast differentiable splatting:
\begin{equation}
    \mathcal{C}_t = (\mathbf{A}_t,\mathbf{R}_t,\mathbf{N}_t)
    = \mathcal{R}_{\mathrm{attr}}(\mathcal{G}_t,\pi_t),
    \label{eq:attribute_rendering}
\end{equation}
where $\mathbf{A}_t$, $\mathbf{R}_t$, and $\mathbf{N}_t$ denote the rendered base color, roughness, and normal maps, respectively, and $\mathcal{R}_{\mathrm{attr}}$ is the differentiable attribute rasterizer.
By projecting the driven 3D proxy into 2D screen space, the rendered buffers $\{\mathcal{C}_t\}$ encode explicit camera and motion control alongside shading-free intrinsic cues, providing dense per-frame guidance for the subsequent neural shader.

\noindent\textbf{Training objectives.}
We train the proxy by supervising the rendered PBR buffers $\{\mathcal{C}_t\}$ at sampled target frames against the intrinsic labels detailed in Sec.~\ref{sec:data_construction}, using L1 ($\mathcal{L}_{\mathrm{L1}}$), SSIM~\cite{wang2004image} ($\mathcal{L}_{\mathrm{SSIM}}$), and LPIPS~\cite{zhang2018unreasonable} ($\mathcal{L}_{\mathrm{LPIPS}}$) losses across all buffers.
Combined with an offset regularizer $\mathcal{L}_{\mathrm{off}}$ that constrains the Gaussians near their FLAME anchors, the overall objective is:
\begin{equation}
    \begin{aligned}
        \mathcal{L}_{\mathrm{off}}
        &= \frac{1}{3N}\sum_{i=1}^{N}
        \|\Delta\boldsymbol{\mu}_i\|_2^2,\\
        \mathcal{L}
        &= \mathcal{L}_{\mathrm{L1}}
        +\lambda_{\mathrm{s}}\mathcal{L}_{\mathrm{SSIM}}+\lambda_{\mathrm{p}}\mathcal{L}_{\mathrm{LPIPS}}
        +\lambda_{\mathrm{off}}\mathcal{L}_{\mathrm{off}},
    \end{aligned}
    \label{eq:stage1_training_loss}
\end{equation}
with $\lambda_{\mathrm{s}}=0.1$, $\lambda_{\mathrm{p}}=0.5$, and $\lambda_{\mathrm{off}}=0.1$.

\subsection{Two-Step Causal Neural Shader}
\label{sec:video_renderer}
Analytical shading of the rasterized PBR buffers $\{\mathcal{C}_t\}$ is sensitive to geometry and material inaccuracies, struggling to reproduce complex light transport and fine facial details.
We therefore formulate neural shading as a real-time causal video generation process, where $\{\mathcal{C}_t\}$ provides view-aligned 3D geometry and intrinsic guidance, allowing the shader to focus on synthesizing photorealistic illumination and high-frequency appearance.

\noindent\textbf{Multi-modal conditioning.}
Built upon Wan2.1-1.3B~\cite{wan2025wan}, our shader adopts TAEHV~\cite{BoerBohan2025TAEHV}, a compact video autoencoder with 2D convolutions and temporal memory blocks, for fast latent encoding and decoding.
It integrates multi-modal conditions according to their structural roles:

\noindent(1) \emph{Per-frame intrinsic and lighting cues}: Each modality in the PBR buffers $\{\mathcal{C}_t\}$ is encoded and concatenated with the noisy video latents along the channel dimension to preserve spatial alignment in latent space.
For dynamic illumination control, the environment map $\mathbf{E}_t$ is represented by LDR, log-HDR, and direction maps, whose encoded features are injected via both channel concatenation and cross-attention.
By associating environment-map texels with fixed 3D directions and jointly conditioning on posed-space normals $\mathbf{N}_t$, the shader learns how incident illumination interacts with surface orientation across motion and viewpoint changes.

\noindent(2) \emph{Global context prefixes}: The static reference $I$ is encoded as a token prefix to anchor subject identity and appearance details across the sequence.
For background harmonization, the target background plate $B$ is encoded as an additional token prefix before the reference tokens to provide scene context, while $\mathbf{E}_t$ is estimated from $B$ at inference via DiffusionLight~\cite{phongthawee2024diffusionlight}.

\noindent\textbf{Causal autoregressive distillation.}
To enable real-time streaming, we formulate the shader as a causal generator that synthesizes video chunks autoregressively in just two denoising steps per chunk, caching keys and values (KV) of the global context prefixes and past chunks.
To inherit broad relighting priors, we distill a pre-trained multi-step bidirectional teacher (Relit-LiVE~\cite{relitlive}) into our streamlined causal student.
Specifically, we first warm up the student on teacher ODE trajectories and then apply distribution matching distillation (DMD)~\cite{yin2024one} over self-forced autoregressive student rollouts~\cite{huang2026self}.
During distillation, the teacher uses its native Wan-VAE to encode intrinsic conditions obtained from DiffusionRenderer~\cite{liang2025diffusionrenderer}, including additional depth and metallic maps, and retains its original joint video--environment interface.
We encode the target log-HDR environment maps with Wan-VAE and perturb both the environment and student-generated video latents to the same noise level, updating the student via DMD using only the video component of the score difference between the frozen teacher and a trainable auxiliary score model.

\noindent\textbf{Paired portrait adaptation.} As the general-purpose teacher lacks domain priors for subtle facial dynamics and skin reflectance, we fine-tune the distilled shader on our paired portrait dataset (Sec.~\ref{sec:data_construction}) via:
\begin{equation}
    \mathcal{L}_{\mathrm{pair}}
    = \lambda_z\mathcal{L}_{\mathrm{latent}}
    + \lambda_{\mathrm{L1}}\mathcal{L}_{\mathrm{L1}}
    + \lambda_{\mathrm{p}}\mathcal{L}_{\mathrm{LPIPS}}
    + \lambda_{\mathrm{adv}}\mathcal{L}_{\mathrm{adv}},
    \label{eq:shader_paired_loss}
\end{equation}
where $\mathcal{L}_{\mathrm{latent}}$ and $\mathcal{L}_{\mathrm{L1}}$ measure latent MSE and RGB $\ell_1$ distance, and $\mathcal{L}_{\mathrm{adv}}$ denotes a patch-based adversarial loss~\cite{isola2017image}.
We set $\lambda_z=\lambda_{\mathrm{L1}}=1$, $\lambda_{\mathrm{p}}=0.5$, and $\lambda_{\mathrm{adv}}=0.01$.
This stage also activates the background prefix stream, unifying environment-map relighting and background harmonization within a single generator via shared illumination conditioning and distilled lighting priors.
More architectural and training details are provided in the supplementary material.

\subsection{Hybrid Relighting Data Curation}
\label{sec:data_construction}
\begin{figure}[t]
    \centering
    \includegraphics[width=\linewidth]{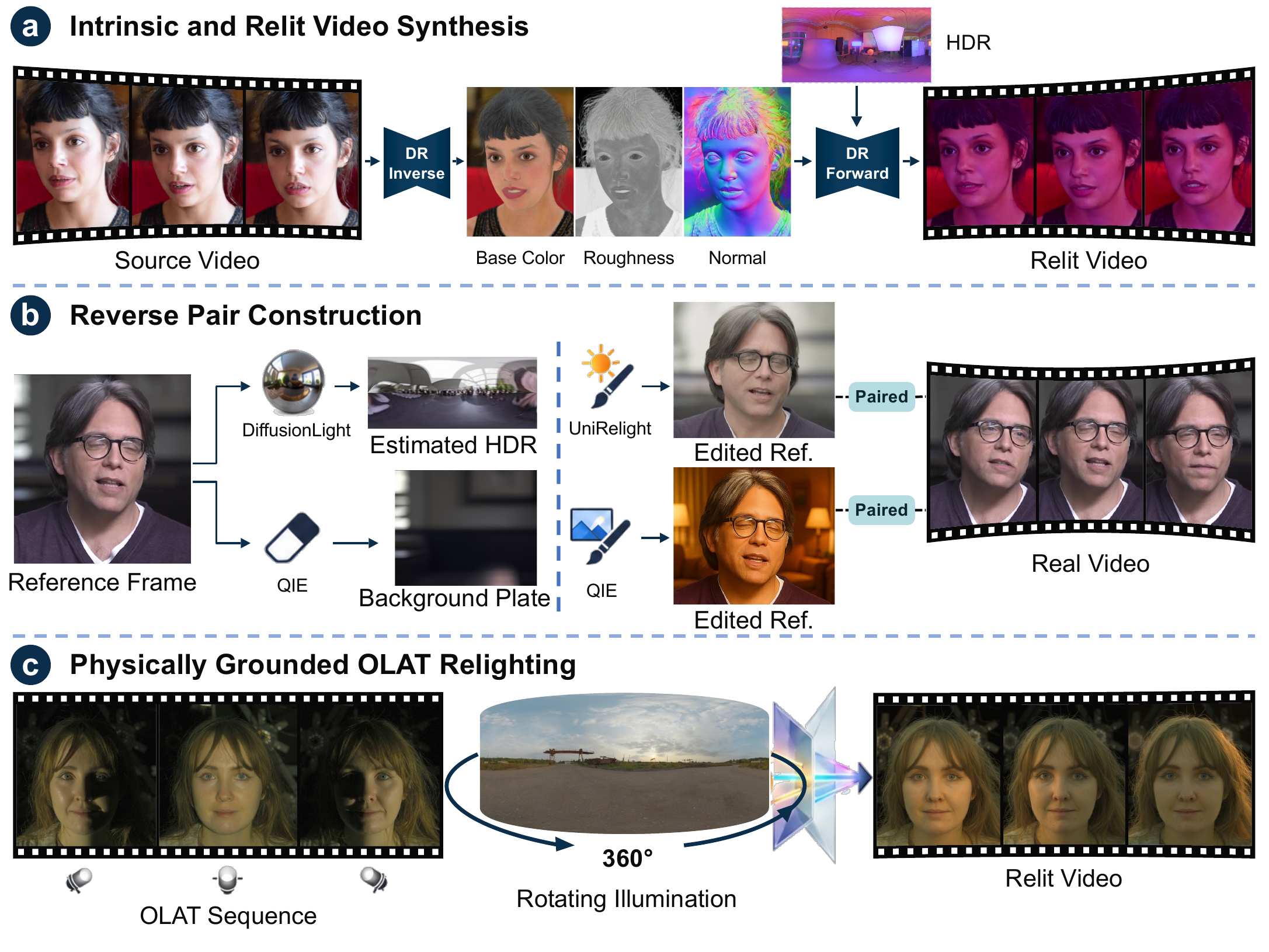}
    \vspace{-3mm}
    \caption{\textbf{Hybrid relighting data curation.} Three complementary data sources provide (a) intrinsic and relit appearance supervision, (b) real video targets preserving facial details and natural motion, and (c) physically grounded supervision under continuously varying illumination.}
    \label{fig:data_construction}
    \vspace{-2mm}
\end{figure}

\begin{table*}[!t]
\centering
\caption{\textbf{Quantitative comparison on the VFHQ and FaceOLAT test sets.}
Best results are in bold.
FPS is based on each method's total inference time for a 57-frame clip.
$^{\dagger}$UIKA + Relit-LiVE uses DiffusionRenderer inverse rendering to obtain PBR inputs before relighting.
$^{*}$With inference optimization on a single GPU.}
\label{tab:main_comparison}
\begingroup
\small
\setlength{\tabcolsep}{1.25pt}
\renewcommand{\arraystretch}{1.05}
\setbox0=\hbox{%
\begin{tabular*}{\textwidth}{@{\extracolsep{\fill}}l*{13}{c}@{}}
\toprule
& \multicolumn{6}{c}{VFHQ} & \multicolumn{6}{c}{FaceOLAT} & \\
\cmidrule(lr){2-7}\cmidrule(lr){8-13}\cmidrule(l){14-14}
Method & PSNR $\uparrow$ & SSIM $\uparrow$ & LPIPS $\downarrow$ & CSIM $\uparrow$ & AED $\downarrow$ & APD $\downarrow$ & PSNR $\uparrow$ & SSIM $\uparrow$ & LPIPS $\downarrow$ & CSIM $\uparrow$ & AED $\downarrow$ & APD $\downarrow$ & FPS $\uparrow$ \\
\midrule
CapHuman & 12.57 & 0.479 & 0.405 & 0.431 & 0.151 & 0.120 & 13.93 & 0.526 & 0.472 & 0.534 & 0.158 & 0.118 & 0.238 \\
ControlFace & 11.70 & 0.497 & 0.367 & 0.506 & 0.188 & 0.117 & 15.80 & 0.539 & 0.411 & 0.637 & 0.225 & 0.115 & 1.132 \\
LCVD & 14.75 & 0.582 & 0.321 & 0.484 & 0.125 & 0.086 & 16.87 & 0.585 & 0.387 & 0.564 & 0.129 & \textbf{0.091} & 0.846 \\
UIKA + UniRelight & 12.56 & 0.529 & 0.337 & 0.511 & 0.136 & 0.105 & 11.89 & 0.537 & 0.438 & 0.503 & 0.198 & 0.166 & 0.135 \\
UIKA + Relit-LiVE$^{\dagger}$ & 13.90 & 0.518 & 0.350 & 0.506 & 0.123 & 0.089 & 15.33 & 0.553 & 0.363 & 0.555 & 0.170 & 0.130 & 0.055 \\
\textbf{Ours} & \textbf{21.32} & \textbf{0.710} & \textbf{0.168} & \textbf{0.739} & \textbf{0.071} & \textbf{0.058} & \textbf{27.83} & \textbf{0.826} & \textbf{0.132} & \textbf{0.747} & \textbf{0.098} & 0.101 & \textbf{12.46 / 20.31}$^{*}$ \\
\bottomrule
\end{tabular*}}
\ifdim\wd0>\textwidth
  \PackageError{main-comparison-table}{Table exceeds text width}{Reduce column spacing.}
\fi
\box0
\par\endgroup
\end{table*}

\begin{figure*}[t]
    \centering
    \includegraphics[width=\linewidth]{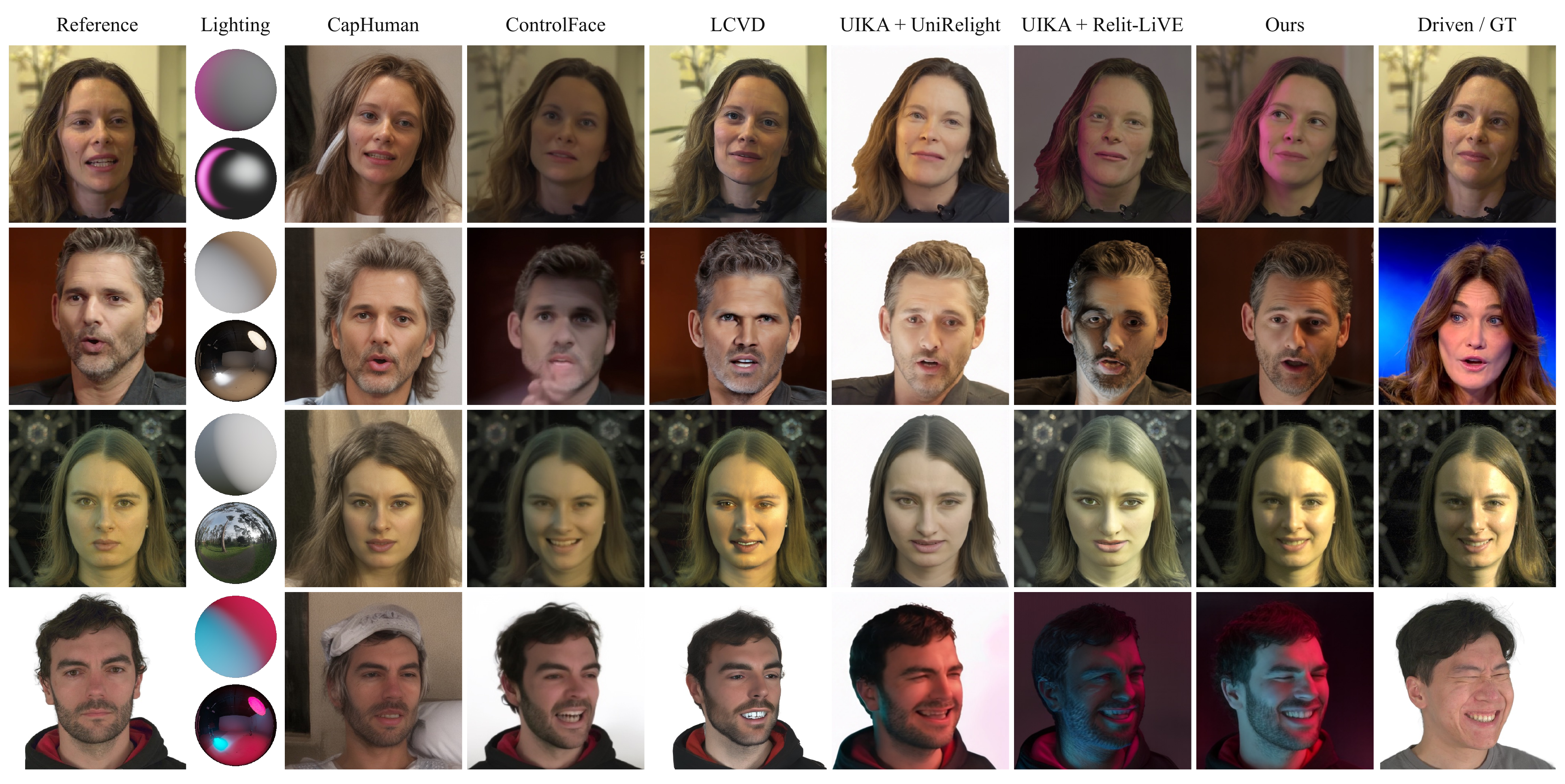}
    \vspace{-3mm}
    \caption{\textbf{Environment-map relighting comparison.}
Rows 1--2 show self- and cross-reenactment on VFHQ, while Rows 3--4 show self-reenactment on FaceOLAT and novel-view cross-reenactment on NeRSemble, respectively.
The last column displays the relit ground truth for Row 3 and driving frames for other rows.
Target illumination is visualized via diffuse (top) and mirror (bottom) spheres.}
    \label{fig:self comparison}
\end{figure*}
Training our framework requires dynamic portrait videos paired with intrinsic buffers and varying illumination, which are prohibitively costly to capture at scale.
As illustrated in Fig.~\ref{fig:data_construction}, we curate three complementary data sources from in-the-wild videos and static OLAT captures to jointly supervise intrinsic decomposition, authentic facial dynamics, and physically accurate light transport.

\noindent\textbf{Intrinsic and relit video synthesis.}
We first leverage DiffusionRenderer~\cite{liang2025diffusionrenderer} to generate both intrinsic labels and relit video pairs.
Its inverse renderer extracts temporally consistent base color, roughness, and normal maps from VFHQ~\cite{xie2022vfhq} to supervise our Gaussian proxy (Sec.~\ref{sec:head_representation}), while its forward renderer relights VFHQ clips under diverse HDR environment maps to provide paired relighting supervision for the shader.

\noindent\textbf{Reverse pair construction via reference editing.}
To prevent the shader from overfitting to synthetic artifacts in synthesized relit targets, we invert the pairing direction by keeping original VFHQ videos as real ground-truth targets while editing sampled reference frames.
For environment-map relighting, we estimate the target HDR illumination from the initial frame via DiffusionLight~\cite{phongthawee2024diffusionlight} and relight the reference frame under random HDR maps using UniRelight~\cite{he2026unirelight}.
For background harmonization, we employ Qwen-Image-Edit~\cite{wu2025qwenimagetechnicalreport} to extract the clean background plate $B$ of the original video and composite the reference portrait onto a newly synthesized background.

\noindent\textbf{Physically grounded OLAT relighting.}
As in-the-wild videos mainly capture facial motion under static lighting, we further introduce continuous illumination changes using public captures from FaceOLAT~\cite{rao20253dpr}.
Exploiting the linearity of light transport, we continuously rotate HDR environment maps over $360^\circ$ and linearly combine the OLAT basis images at each step, yielding physically accurate relit sequences under dynamic lighting.

\section{Experiment}
\label{sec:experiment}
\subsection{Experimental Setup}
\label{sec:experimental_setup}
\noindent\textbf{Implementation details.}
We train our framework in two stages.
First, the intrinsic Gaussian head proxy uses a frozen DINOv3 ViT-L/16 backbone with a $48\times48$ UV query grid to predict $192\times192$ canonical attribute maps.
It is trained on $512\times512$ images for 100K iterations using AdamW with a learning rate of $4\times10^{-4}$ and a batch size of 64.
Second, the causal video shader is distilled via 3K ODE regression and 2K DMD steps, and then fine-tuned on our hybrid dataset for 5K iterations across 32 NVIDIA H20 GPUs (batch size 32) using Muon ($10^{-4}$ learning rate) and auxiliary AdamW updates ($10^{-5}$).
Paired adaptation uses 57-frame clips for synthesized relit videos and OLAT sequences, and 81-frame clips for reverse-paired data.
At inference, we reconstruct the canonical proxy once and autoregressively generate chunks of three latent frames using two denoising steps per chunk with KV-cache reuse.
To bound the attention context during streaming, self-attention accesses the current and preceding chunks together with up to two uniformly selected chunks from earlier history.
More details are provided in the supplementary material.

\noindent\textbf{Datasets.}
Public benchmarks for head animation under dynamic illumination remain limited.
Following Sec.~\ref{sec:data_construction}, we construct training and held-out test sets from VFHQ~\cite{xie2022vfhq} and FaceOLAT~\cite{rao20253dpr} for qualitative and quantitative evaluation.
To assess cross-dataset generalization, we further conduct qualitative comparisons on NeRSemble~\cite{kirschstein2023nersemble}, which is unseen during training.

\noindent\textbf{Baselines.}
We primarily evaluate single-portrait head animation jointly with relighting or background harmonization.
For environment-map relighting, we compare against 2D generative models CapHuman~\cite{liang2024caphuman}, ControlFace~\cite{jang2025controlface}, and LCVD~\cite{guo2025high}, and construct cascaded baselines by pairing the 3D head animator UIKA~\cite{wu2026uika} with the video relighting models UniRelight~\cite{he2026unirelight} or Relit-LiVE~\cite{relitlive}.
For background harmonization, we similarly pair the 2D animator Follow-Your-Emoji (FYE)~\cite{ma2026follow} or UIKA with the video harmonization models RelightVid~\cite{fang2025relightvid} or UniLumos~\cite{liu2026unilumos}.
We omit comparisons with RelightAnyone~\cite{xu2026relightanyone} and URAvatar~\cite{li2024uravatar} due to the lack of public implementations.

\noindent\textbf{Evaluation metrics.}
We measure foreground reconstruction quality via PSNR, SSIM, and LPIPS, identity preservation via ArcFace~\cite{deng2019arcface} cosine similarity (CSIM), and motion accuracy via average expression and pose distances (AED/APD).
CSIM is averaged over frames with valid face detections across all methods.
We benchmark inference throughput in FPS on a single NVIDIA RTX 5090 D GPU.
\subsection{Comparisons with Existing Methods}
\label{sec:comparisons}

\begin{figure}[t]
    \centering
    \includegraphics[width=\linewidth]{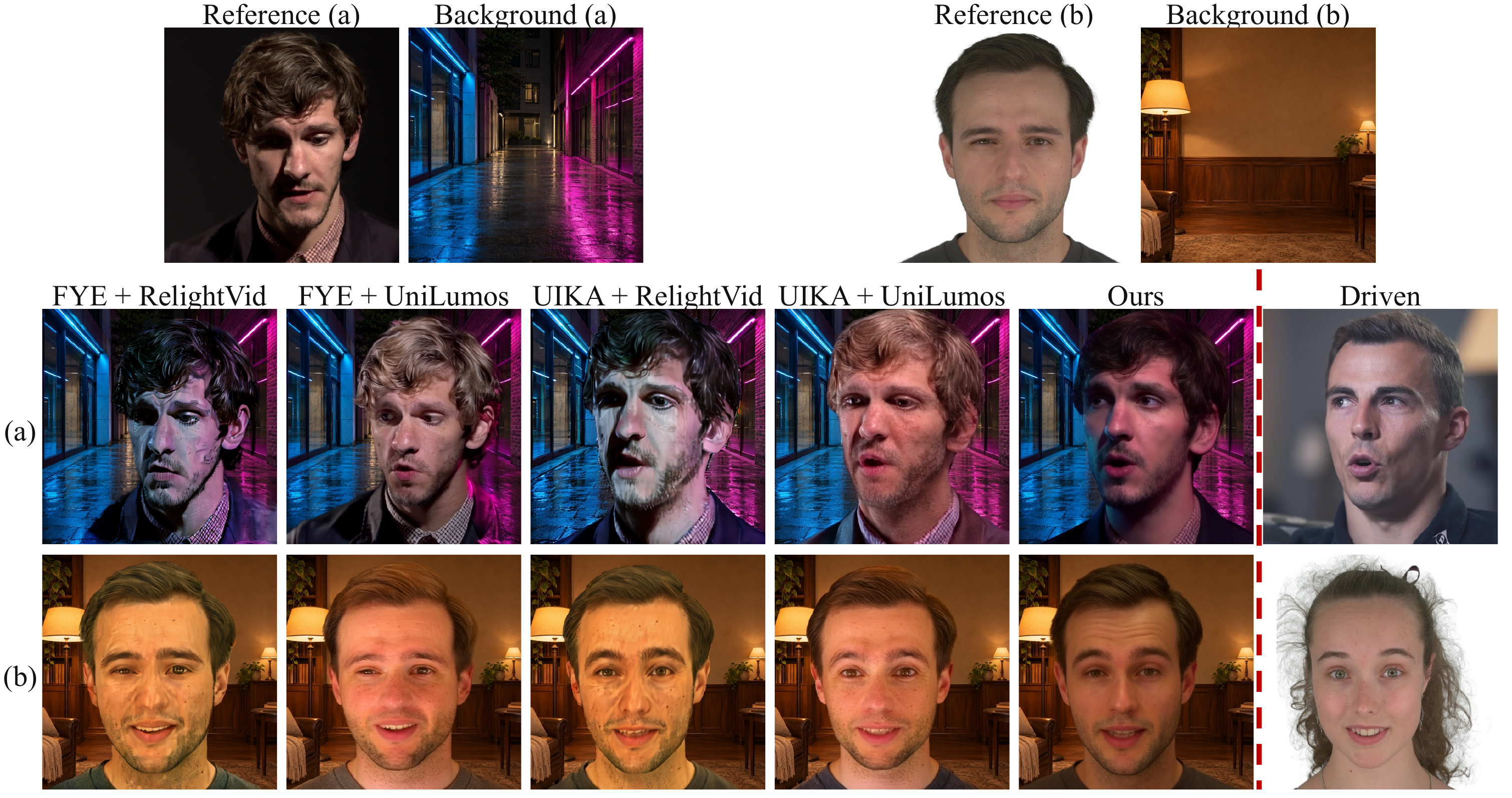}
    \vspace{-3mm}
    \caption{\textbf{Background harmonization comparison.}
We show cross-reenactment results on (a) VFHQ and (b) NeRSemble, with reference images and target backgrounds shown at the top.
The last column displays the driving frames.}
    \label{fig:bg_harmonization}
    \vspace{-2mm}
\end{figure}

\noindent\textbf{Environment-map relighting.}
As shown in Fig.~\ref{fig:self comparison}, CapHuman, ControlFace and LCVD struggle with identity preservation and lighting control.
UIKA-based cascades inherit upstream deformation artifacts, as downstream video relighters are trained to preserve input structure.
Moreover, their general-scene priors often yield less convincing facial shading.
In contrast, our full pipeline achieves more faithful lighting responses with superior identity preservation and reconstruction quality, consistent with the metrics in Table~\ref{tab:main_comparison}.
The supplementary material provides additional qualitative results for the above comparisons, as well as evaluations of direct relighting methods (3DPR~\cite{rao20253dpr}, Lite2Relight~\cite{rao2024lite2relight}, DiffusionRenderer~\cite{liang2025diffusionrenderer}, Relit-LiVE, and UniRelight) using target-pose RGB frames as input.

\noindent\textbf{Background harmonization.}
Figure~\ref{fig:bg_harmonization} compares harmonization results on backgrounds synthesized by GPT Image 2.
Under challenging illumination, both 2D and 3D cascades paired with RelightVid or UniLumos exhibit appearance distortions or unnatural shading.
By sharing a common relighting prior across environment-map relighting and background harmonization, our method produces more realistic facial appearance and natural lighting effects, with quantitative results reported in the supplementary material.

Furthermore, our method runs substantially faster than baselines, achieving 20.31\,FPS on a single GPU and 26.02\,FPS on two GPUs with inference optimization.
Finally, since predicted target illumination on VFHQ may deviate from the true lighting, quantitative metrics alone cannot fully reflect perceptual quality.
We thus provide a user study in the supplementary material.

\subsection{Ablation Studies}
\label{sec:ablation}

\begin{table}[t]
\centering
\caption{\textbf{Training strategy and intrinsic proxy ablations.}
}
\label{tab:training_proxy_ablation}
\begingroup
\footnotesize
\setlength{\tabcolsep}{2pt}
\renewcommand{\arraystretch}{1.0}
\begin{tabular*}{\linewidth}{@{\extracolsep{\fill}}l*{6}{c}@{}}
\toprule
& \multicolumn{3}{c}{VFHQ} & \multicolumn{3}{c}{FaceOLAT} \\
\cmidrule(lr){2-4}\cmidrule(l){5-7}
Variant & PSNR $\uparrow$ & SSIM $\uparrow$ & LPIPS $\downarrow$ & PSNR $\uparrow$ & SSIM $\uparrow$ & LPIPS $\downarrow$ \\
\midrule
DMD-only & 16.38 & 0.625 & 0.298 & 16.41 & 0.652 & 0.414 \\
Direct SF & 19.21 & 0.663 & 0.217 & 23.95 & 0.775 & 0.192 \\
\midrule
GS-RGB & 18.84 & 0.652 & 0.238 & 25.64 & 0.806 & 0.155 \\
GS-N & 20.68 & 0.697 & 0.182 & 26.66 & 0.812 & 0.147 \\
FLAME-N & 20.15 & 0.687 & 0.192 & 26.38 & 0.821 & 0.142 \\
\midrule
\textbf{Ours} & \textbf{21.32} & \textbf{0.710} & \textbf{0.168} & \textbf{27.83} & \textbf{0.826} & \textbf{0.132} \\
\bottomrule
\end{tabular*}
\par\endgroup
\end{table}

\noindent\textbf{Effect of training strategy and intrinsic proxy.}
Table~\ref{tab:training_proxy_ablation} ablates our training strategy and intrinsic proxy on the VFHQ and FaceOLAT test sets.

\emph{Training strategy.}
We compare our full training pipeline against two variants: (1) DMD-only, which evaluates the causal shader after 2K distillation steps without paired portrait adaptation, and (2) Direct SF, which fine-tunes a generic Self-Forcing initialization on our paired data, bypassing relighting distillation.
Our full model consistently outperforms both variants across all metrics, confirming that distilling relighting priors and adapting to paired portrait data are both essential and complementary.

\emph{Intrinsic proxy.}
We replace our full PBR buffers $\{\mathcal{C}_t\}$ with three alternatives while keeping all other conditions unchanged: (1) GS-RGB, which renders standard RGB Gaussians with baked-in source lighting; (2) GS-N, which provides only our Gaussian normal maps $\{\mathbf{N}_t\}$; and (3) FLAME-N, which rasterizes coarse FLAME mesh normals.
GS-N and FLAME-N show mixed results on FaceOLAT, whose static portraits do not fully reflect the benefits of animated normals.
These benefits are more evident on VFHQ, where GS-N outperforms FLAME-N across all metrics.
Conditioning on the full PBR buffers consistently achieves the best results across all datasets, demonstrating the necessity of shading-free material cues $\{\mathbf{A}_t, \mathbf{R}_t\}$ and fine-grained geometry $\{\mathbf{N}_t\}$.
\begin{table}[t]
\centering
\caption{\textbf{Long-sequence stability.}
Evaluated on five sequences of 982--984 frames. FG/BG denote foreground/background.}
\label{tab:long_sequence_stability}
\begingroup
\footnotesize
\setlength{\tabcolsep}{3.5pt}
\renewcommand{\arraystretch}{1.0}
\setbox0=\hbox{%
\begin{tabular*}{\linewidth}{@{\extracolsep{\fill}}rccccc@{}}
\toprule
Rollout & \multicolumn{3}{c}{Full image} & \multicolumn{2}{c}{PSNR $\uparrow$} \\
\cmidrule(lr){2-4}\cmidrule(l){5-6}
(frames) & PSNR $\uparrow$ & SSIM $\uparrow$ & LPIPS $\downarrow$ & FG & BG \\
\midrule
81 & 21.00 & 0.7062 & 0.2165 & 21.26 & 22.22 \\
333 & 20.65 & 0.7055 & 0.2218 & 21.02 & 21.55 \\
Full & 20.50 & 0.7039 & 0.2277 & 20.99 & 21.25 \\
\bottomrule
\end{tabular*}}
\ifdim\wd0>\linewidth
  \PackageError{long-sequence-table}{Table exceeds column width}{Reduce column spacing.}
\fi
\box0
\par\endgroup
\end{table}

\begin{figure}[t]
    \centering
    \includegraphics[width=0.85\linewidth]{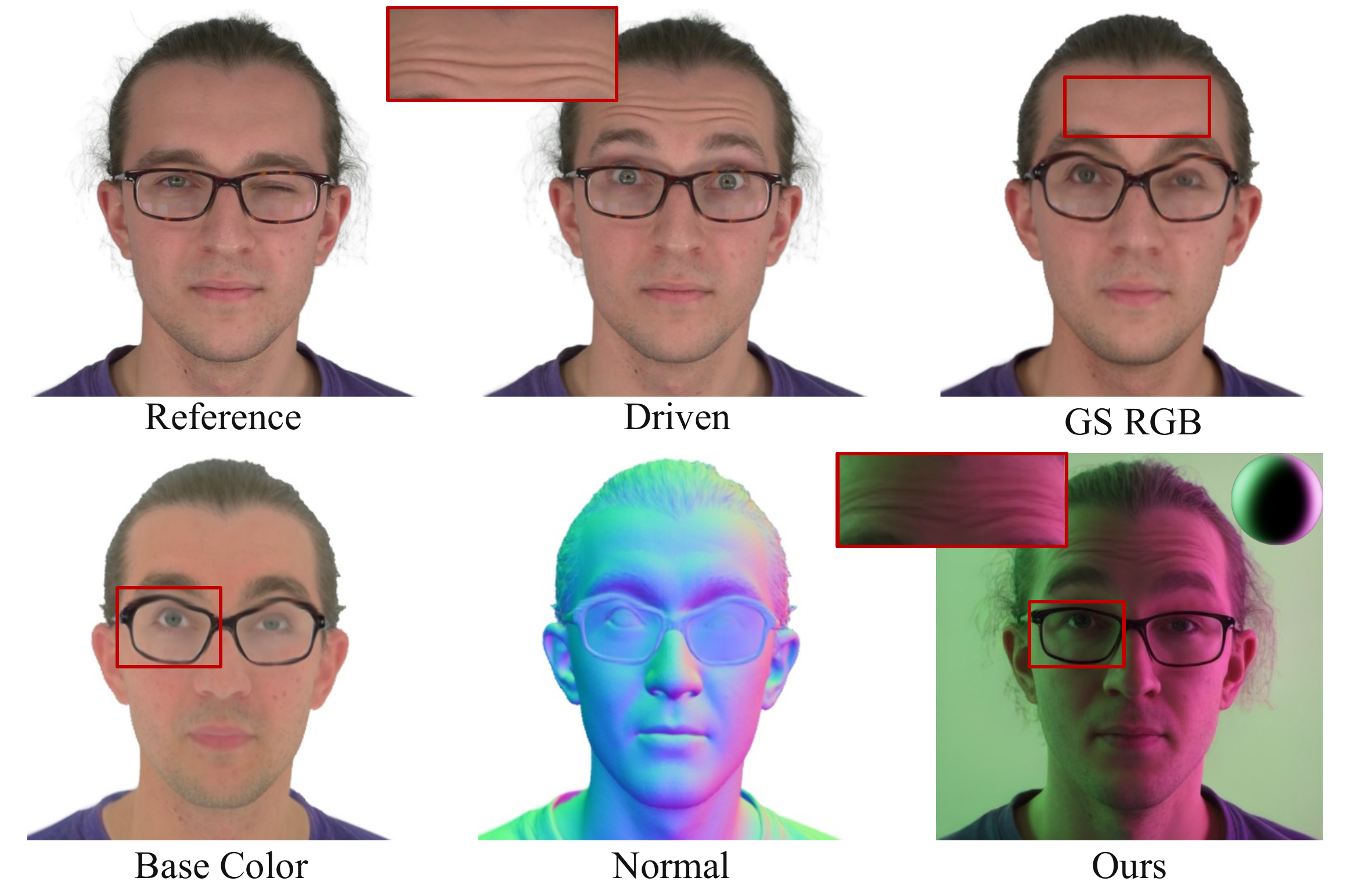}
    \caption{\textbf{Effect of the causal neural shader.}
On the unseen NeRSemble dataset, our shader synthesizes expression-dependent wrinkles and reduces eyeglass distortion.}
    \label{fig:gs ablation}
    \vspace{-2mm}
\end{figure}

\noindent\textbf{Effect of the causal neural shader.}
On the unseen NeRSemble dataset (Fig.~\ref{fig:gs ablation}), the LBS-driven Gaussian proxy fails to produce expression-dependent forehead wrinkles missing from the reference image and introduces non-rigid bending along the eyeglass rims.
Our neural shader leverages learned video priors to synthesize realistic dynamic wrinkles and preserve the structural rigidity of the glasses under novel illumination, generalizing effectively to unseen domains without per-subject optimization.

\begin{table}[t]
\caption{\textbf{Runtime ablation on RTX 5090 D GPUs.}
(a) Single-GPU causal neural shader ablation. (b) Optimized dual-GPU full-pipeline breakdown.}
\label{tab:speed ablation}
\centering
\begingroup
\footnotesize
\setlength{\tabcolsep}{2pt}
\renewcommand{\arraystretch}{1.02}
\newcommand{\SpeedRowGap}{0pt}
\newcommand{\SpeedTableLeft}{%
\begin{tabular}{@{}lrr@{}}
\toprule
Method & Time (s) & FPS \\
\midrule
Relit-LiVE & 556.51 & 0.10 \\[\SpeedRowGap]
w/ Wan-VAE & 42.24 & 1.35 \\[\SpeedRowGap]
w/ token concat. & 24.06 & 2.37 \\[\SpeedRowGap]
w/ both  & 77.92 & 0.73 \\[\SpeedRowGap]
\midrule
\textbf{Ours} & \textbf{4.19} & \textbf{13.60} \\[\SpeedRowGap]
\textbf{Ours + optimization} & \textbf{2.42} & \textbf{23.55} \\
\bottomrule
\end{tabular}}
\setbox0=\hbox{\SpeedTableLeft}
\setbox2=\hbox{%
\begin{tabular}{@{}lr@{}}
\toprule
Block & Time (s) \\
\midrule
\multicolumn{2}{@{}l@{}}{\fontsize{5.5}{6}\selectfont\itshape GPU~1}\\[-2pt]
2-step denoising & 32.61 \\
\midrule
\multicolumn{2}{@{}l@{}}{\fontsize{5.5}{6}\selectfont\itshape GPU~2}\\[-2pt]
GS recon.  & 0.14 \\
GS rendering & 2.60 \\
Cond. encoding & 10.23 \\
RGB decoding & 2.10 \\
\midrule
GPU transfer & 0.68 \\
\midrule
\textbf{Total / FPS} & \textbf{37.82 / 26.02} \\
\bottomrule
\end{tabular}}
\dimen0=\dimexpr\ht2+\dp2-\ht0-\dp0\relax
\ifdim\dimen0>0pt
  \divide\dimen0 by 5
  \edef\SpeedRowGap{\the\dimen0}
  \setbox0=\hbox{\SpeedTableLeft}
\fi
\dimen0=\dimexpr\ht2+\dp2-\ht0-\dp0\relax
\ifdim\dimen0<0pt \dimen0=-\dimen0\fi
\ifdim\dimen0>0.1pt
  \PackageError{speed-table}{Panel heights do not match}{Check the left row gaps.}
\fi
\begin{minipage}[t]{0.55\linewidth}
\centering\vspace{0pt}
\textbf{(a) Causal neural shader}\\[-1pt]
{\scriptsize 1 GPU, 57 frames}\\[3pt]
\ifdim\wd0>\linewidth
  \PackageError{speed-table}{Left panel exceeds allocated width}{Reduce table width.}
\fi
\box0
\end{minipage}\hfill%
\begin{minipage}[t]{0.43\linewidth}
\centering\vspace{0pt}
\textbf{(b) Full pipeline}\\[-1pt]
{\scriptsize 2 GPUs, 984 frames}\\[3pt]
\ifdim\wd2>\linewidth
  \PackageError{speed-table}{Right panel exceeds allocated width}{Reduce table width.}
\fi
\box2
\end{minipage}
\par\endgroup
\end{table}

\begin{figure}[t]
    \centering
    \includegraphics[width=\linewidth]{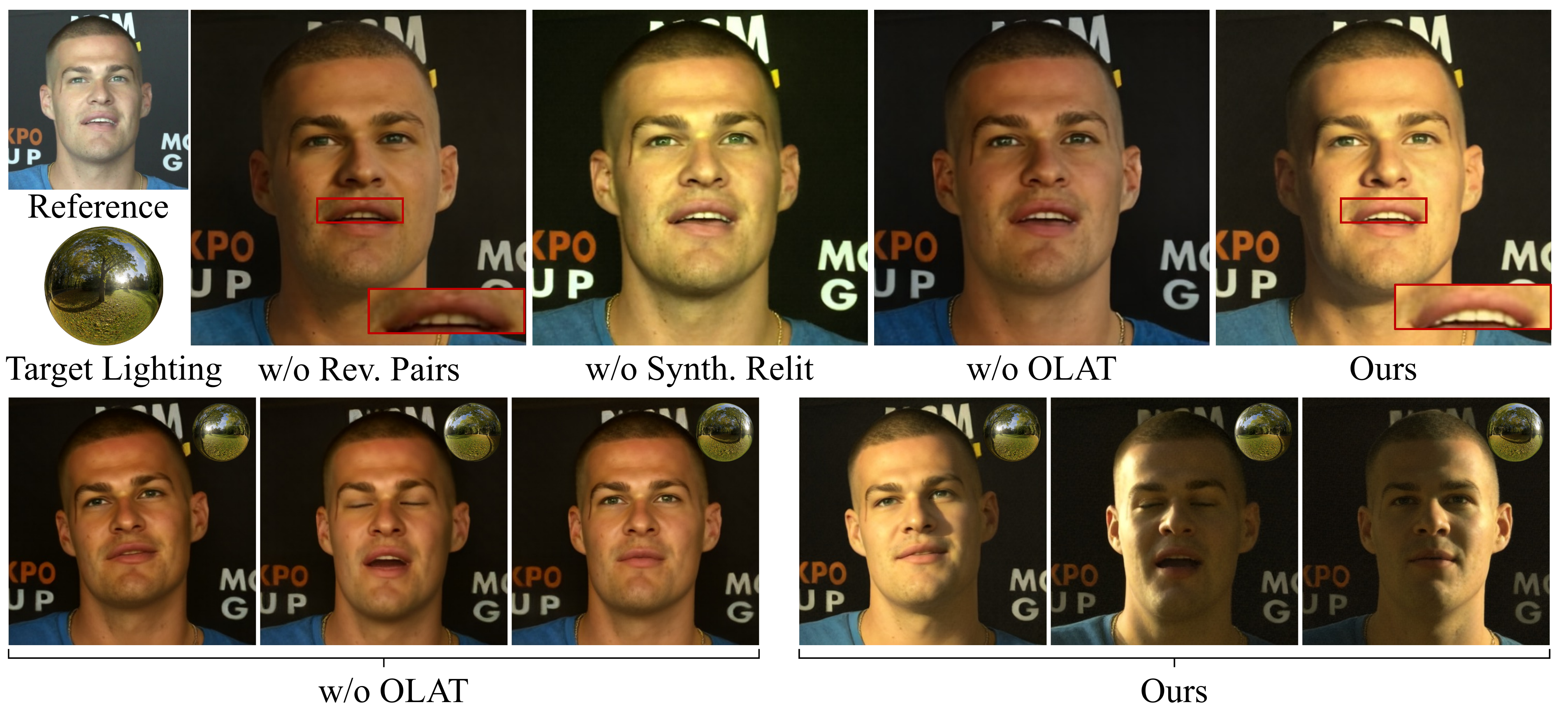}
    \caption{\textbf{Effect of hybrid training data.}
Top: relighting comparison under the same target lighting.
Bottom: head animation under dynamic lighting.}
    \label{fig:data ablation}
    \vspace{-2mm}
\end{figure}

\noindent\textbf{Effect of hybrid training data.}
Fig.~\ref{fig:data ablation} demonstrates the complementary roles of our three training data sources.
While synthesized relit videos from DiffusionRenderer provide diverse illumination supervision, removing reverse pairs visibly degrades mouth and tooth details, as real video targets are crucial for preserving high-frequency appearance and faithful motion adherence.
Furthermore, incorporating static OLAT captures under rotating illumination provides physically grounded supervision for continuous lighting transitions, significantly improving dynamic shading fidelity on moving heads without requiring costly dynamic light stage captures.

\noindent\textbf{Long-sequence stability.}
To assess error accumulation during streaming, we evaluate five VFHQ test sequences of 982--984 frames at rollout lengths of 81, 333, and the full sequence.
Shorter rollouts periodically reset the inference state between segments, whereas the full rollout streams continuously without resetting the KV cache.
Table~\ref{tab:long_sequence_stability} shows minimal foreground degradation, with an average PSNR drop of only 0.27\,dB from 81-frame to full rollouts, while the background drift largely stems from inherent camera jitter in the ground-truth VFHQ videos.

\noindent\textbf{Runtime ablation.}
Table~\ref{tab:speed ablation}(a) evaluates our standalone shader over 57 frames, excluding Gaussian reconstruction and rasterization.
Our two-step shader takes only 4.19\,s versus 556.51\,s for the 50-step Relit-LiVE ($132.82\times$ speedup), while outperforming two-step variants with Wan-VAE and token concatenation by up to $18.60\times$.
Inference optimization, detailed in the supplementary material, further cuts runtime to 2.42\,s at 23.55\,FPS.
Table~\ref{tab:speed ablation}(b) evaluates the end-to-end dual-GPU pipeline over 984 frames, where GPU~1 performs two-step denoising while GPU~2 concurrently executes Gaussian reconstruction, rasterization, condition encoding, and latent decoding, achieving 26.02\,FPS.

\section{Conclusion}
\label{sec:conclusion}
We present LuxAvatar, a feed-forward framework for real-time head avatar animation and relighting from a single portrait.
By coupling an animatable 3D intrinsic Gaussian proxy with a causal neural shader, our approach leverages view- and motion-consistent PBR buffers as 4D guidance to synthesize fine-scale appearance and complex light transport beyond coarse 3D representations.
Heterogeneous distillation further enables low-latency autoregressive streaming, while our hybrid training strategy adapts general relighting priors to authentic facial dynamics and physically grounded lighting changes.
Extensive experiments demonstrate temporally coherent, real-time head animation, relighting, and background harmonization, without requiring per-subject optimization or dynamic light stage captures.


\section*{Acknowledgments}
We thank Zijian Wu, Qiuyu Wang, and Hao Ouyang for valuable discussions.

{
    \small
    \bibliographystyle{ieeenat_fullname}
    \bibliography{main}
}

\clearpage
\setcounter{page}{1}
\setcounter{section}{0}
\setcounter{equation}{0}
\setcounter{table}{0}
\setcounter{figure}{0}
\renewcommand{\thesection}{\Alph{section}}
\renewcommand{\theequation}{S\arabic{equation}}
\renewcommand{\thetable}{S\arabic{table}}
\renewcommand{\thefigure}{S\arabic{figure}}
\maketitlesupplementary

This supplementary material provides architectural, training, and data-curation details (Sec.~\ref{sec:supp_implementation}), streaming and inference optimizations (Sec.~\ref{sec:supp_streaming}), additional evaluations (Sec.~\ref{sec:supp_evaluation}), and limitations (Sec.~\ref{sec:supp_limitations}).

\section{Architecture and Training Details}
\label{sec:supp_implementation}

\subsection{Intrinsic Gaussian Head Proxy}
\label{sec:supp_proxy}
\noindent\textbf{Architecture and attribute parameterization.}
In our canonical proxy, image-space and UV-space DINOv3 ViT-L/16~\cite{simeoni2025dinov3} token streams are fused by a 4-layer transformer ($1024$-dim tokens) using $48\times48$ learnable UV queries, and upsampled by the DPT-style decoder~\cite{ranftl2021vision} into a $192\times192\times128$ canonical feature map $\mathbf{F}$.
Gaussians are instantiated exclusively at texels within the valid FLAME UV mask~\cite{FLAME:SiggraphAsia2017}, with base color $\mathbf{a}_i$ and roughness $r_i$ bounded to $[0,1]$ via sigmoid activations and canonical normals $\mathbf{n}_i$ normalized to unit length.

\noindent\textbf{Deformation and attribute splatting.}
At frame $t$, we deform the canonical Gaussians using FLAME~\cite{FLAME:SiggraphAsia2017} expression coefficients $\boldsymbol{\psi}_t$, joint poses, and global translation $\mathbf{t}_t$.
The expression basis $\mathbf{B}_{\mathrm{expr}}(\mathbf{u}_i)$ and blended skinning transform are barycentrically interpolated at each Gaussian's UV coordinate $\mathbf{u}_i$.
Writing the latter's linear and translation components as $\mathbf{M}_{i,t}\in\mathbb{R}^{3\times3}$ and $\mathbf{d}_{i,t}\in\mathbb{R}^{3}$, respectively, we obtain the posed center $\boldsymbol{\mu}_{i,t}$ and unit normal $\mathbf{n}_{i,t}$ as follows:
\begin{equation}
\begin{aligned}
    \boldsymbol{\mu}_{i,t}
    &= \mathbf{M}_{i,t}\!\left(\boldsymbol{\mu}_i+\mathbf{B}_{\mathrm{expr}}(\mathbf{u}_i)\boldsymbol{\psi}_t\right)
    +\mathbf{d}_{i,t}+\mathbf{t}_t,\\
    \mathbf{n}_{i,t}
    &= \frac{\mathbf{M}_{i,t}\mathbf{n}_i}{\|\mathbf{M}_{i,t}\mathbf{n}_i\|_2},
\end{aligned}
\label{eq:supp_lbs}
\end{equation}
while $(\mathbf{s}_i,\mathbf{q}_i,\alpha_i,\mathbf{a}_i,r_i)$ remain shared across frames.
Under the target camera $\pi_t$, we render each PBR buffer using 3DGS~\cite{kerbl20233d}.
Its value $\mathbf{C}_t(\mathbf{x})$ at pixel $\mathbf{x}$ is obtained by alpha-compositing the corresponding Gaussian attributes $\mathbf{c}_{i,t}$ from front to back:
\begin{equation}
    \mathbf{C}_t(\mathbf{x})
    = \sum_{i\in\mathcal{N}(\mathbf{x})}
    \tilde{\alpha}_{i,t}(\mathbf{x})T_{i,t}(\mathbf{x})\,\mathbf{c}_{i,t}
    + T_{\mathrm{bg},t}(\mathbf{x})\,\mathbf{b}.
\label{eq:supp_splatting}
\end{equation}
Here, $\mathcal{N}(\mathbf{x})$ indexes the contributing Gaussians in depth order, and $\tilde{\alpha}_{i,t}(\mathbf{x})$ is the projected opacity of Gaussian $i$.
The transmittance before Gaussian $i$ is $T_{i,t}(\mathbf{x})=\prod_{j<i}(1-\tilde{\alpha}_{j,t}(\mathbf{x}))$, where $j<i$ follows this ordering.
The remaining transmittance $T_{\mathrm{bg},t}(\mathbf{x})=\prod_{i\in\mathcal{N}(\mathbf{x})}(1-\tilde{\alpha}_{i,t}(\mathbf{x}))$ weights the constant background value $\mathbf{b}$, fixed to white when rendering PBR conditions for the shader.
The target background plate $B$ is provided separately to the neural shader and is not modeled by the Gaussian proxy.

\subsection{Causal Neural Shader}
\label{sec:supp_shader}
\noindent\textbf{Latent grid and multi-modal conditioning.}
The frozen TAEHV autoencoder~\cite{BoerBohan2025TAEHV} applies $8\times$ spatial downsampling and $4\times$ temporal compression into a 16-channel latent space (mapping 57- and 81-frame RGB clips to 15 and 21 latent frames), which our Wan2.1-1.3B~\cite{wan2025wan} backbone generates autoregressively in 3-frame latent chunks.
In the channel-concatenation branch, the three PBR buffers and three illumination maps (LDR, log-HDR, and 3D direction maps) are independently encoded at the input video scale into six spatially aligned 16-channel latents and concatenated with the 16-channel noisy video latent, forming a 112-channel patch-embedding input without increasing the video token count.
In the parallel environment cross-attention branch, the three lighting maps are spatially downsampled before $8\times$ autoencoder compression and $2\times$ strided projection into compact lighting context tokens concatenated with a fixed empty-prompt embedding from the frozen text encoder.
Reference and background latents are separately projected into spatial token grids with the same feature dimension as video tokens, increasing sequence length without changing the 112-channel video input.
With a background prefix, BG and reference occupy the first two temporal positions, and video RoPE indices and KV-cache offsets are shifted accordingly.
Chunk-causal attention allows video tokens to access both prefixes and available video history, while excluding future chunks.

\begin{table*}[t]
\centering
\caption{\textbf{Teacher and student interfaces.}}
\label{tab:supp_shader_interfaces}
\small
\begin{tabular*}{\textwidth}{@{\extracolsep{\fill}}lll@{}}
\toprule
Component & Teacher / auxiliary score model & Causal student \\
\midrule
Condition autoencoder & Wan-VAE & TAEHV \\
Intrinsic modalities & Base color, roughness, normal, depth, metallic & Base color, roughness, normal \\
Intrinsic source & DiffusionRenderer & Intrinsic Gaussian head proxy \\
Lighting modalities & LDR, log-HDR, directions & LDR, log-HDR, directions \\
Temporal attention & Bidirectional & Chunk-causal \\
Prediction interface & Joint video--environment & Video only \\
Role during DMD & Frozen teacher / trainable auxiliary model & Trainable generator \\
\bottomrule
\end{tabular*}
\end{table*}

\begin{figure*}[t]
    \centering
    \includegraphics[width=\textwidth]{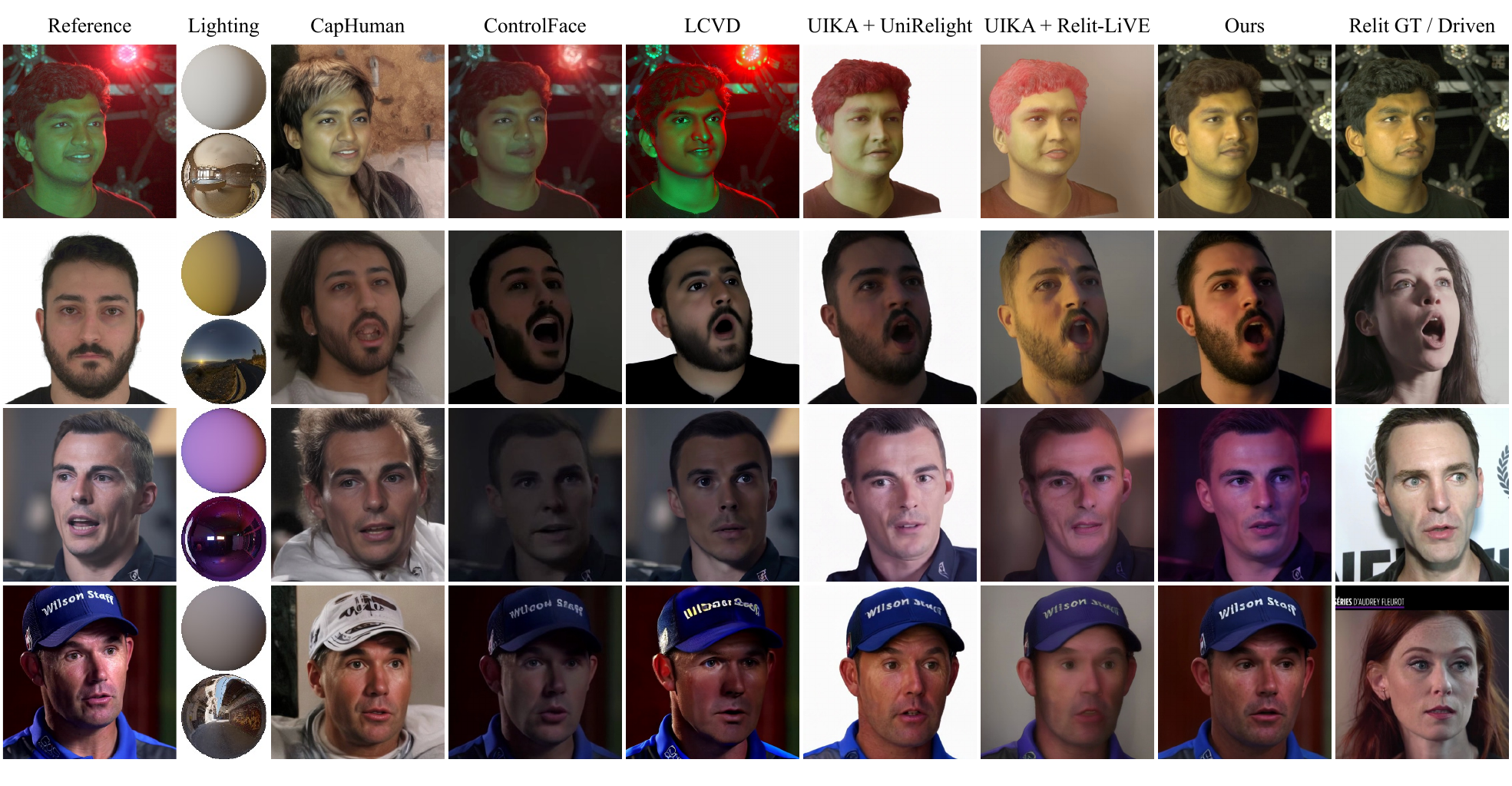}
    \caption{\textbf{Additional environment-map relighting comparisons.}
The last column shows relit ground truth in the first row and driving frames in the remaining rows.
Target lighting is visualized by diffuse (top) and mirror (bottom) spheres.}
    \label{fig:supp_env_relighting}
\end{figure*}

\begin{figure*}[t]
    \centering
    \includegraphics[width=\textwidth]{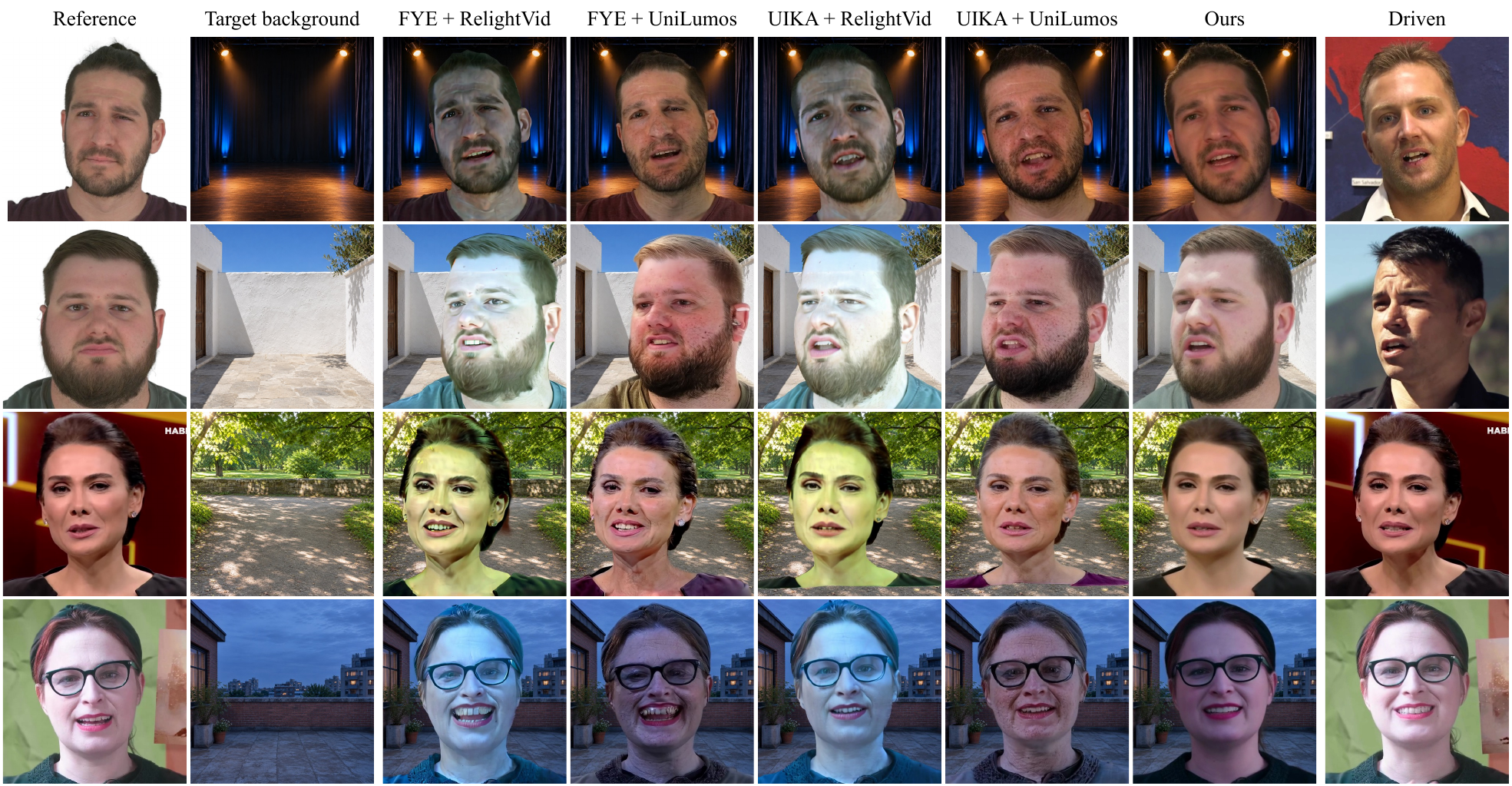}
    \caption{\textbf{Additional background harmonization comparisons.}
The first two columns show the reference portraits and target backgrounds.
The last column shows the driving frames.}
    \label{fig:supp_background_harmonization}
\end{figure*}

\begin{figure*}[p]
    \begin{minipage}[c][\textheight][c]{\textwidth}
    \centering
    \includegraphics[width=\textwidth,height=0.94\textheight,keepaspectratio]{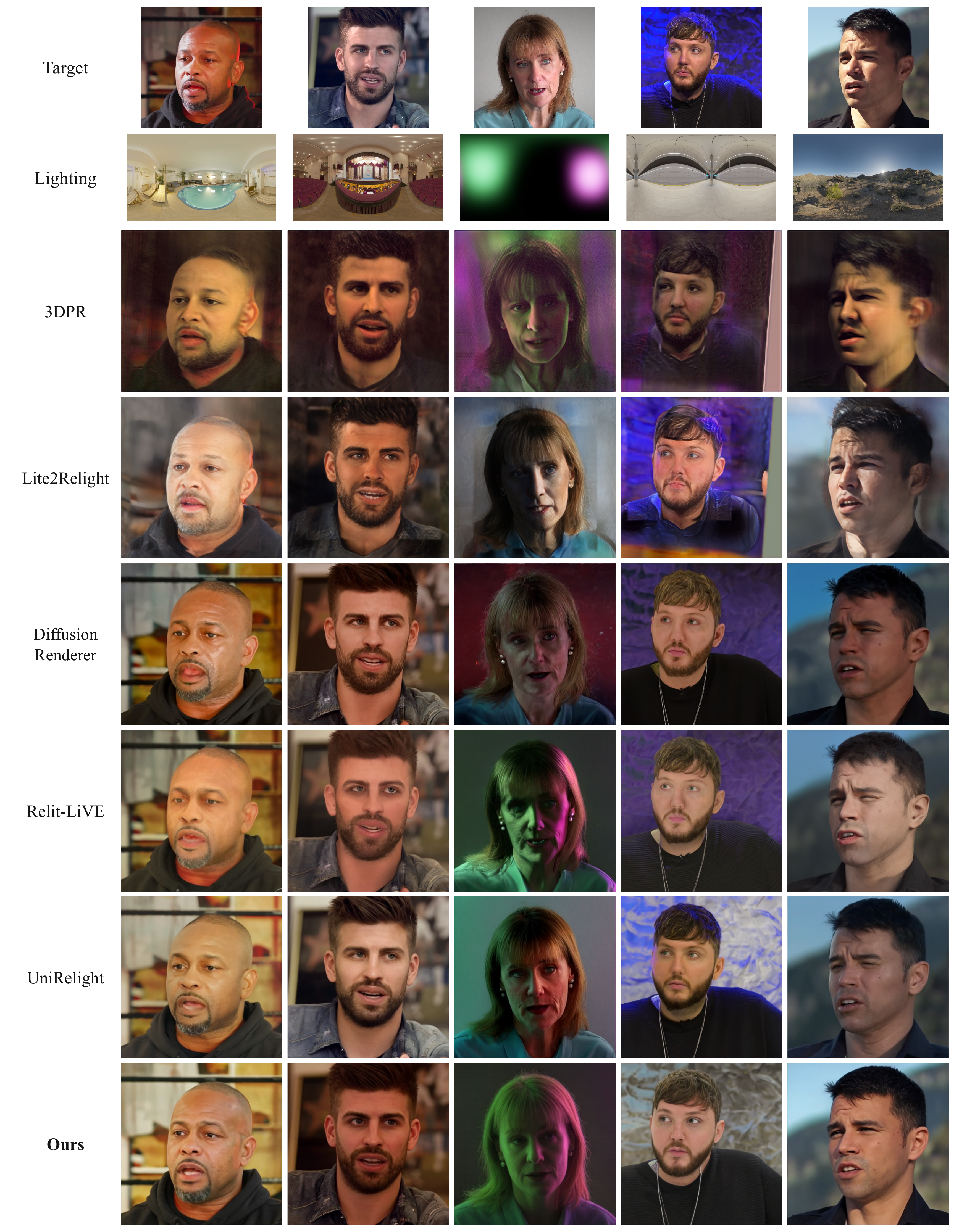}
    \caption{\textbf{Qualitative comparison with direct relighting methods.}
The first two rows show target-pose RGB frames and target environment maps, respectively.
Subsequent rows show the results of each method.}
    \label{fig:supp_direct_relighting}
    \end{minipage}
\end{figure*}

\begin{figure}[t]
\centering
    \includegraphics[width=\linewidth]{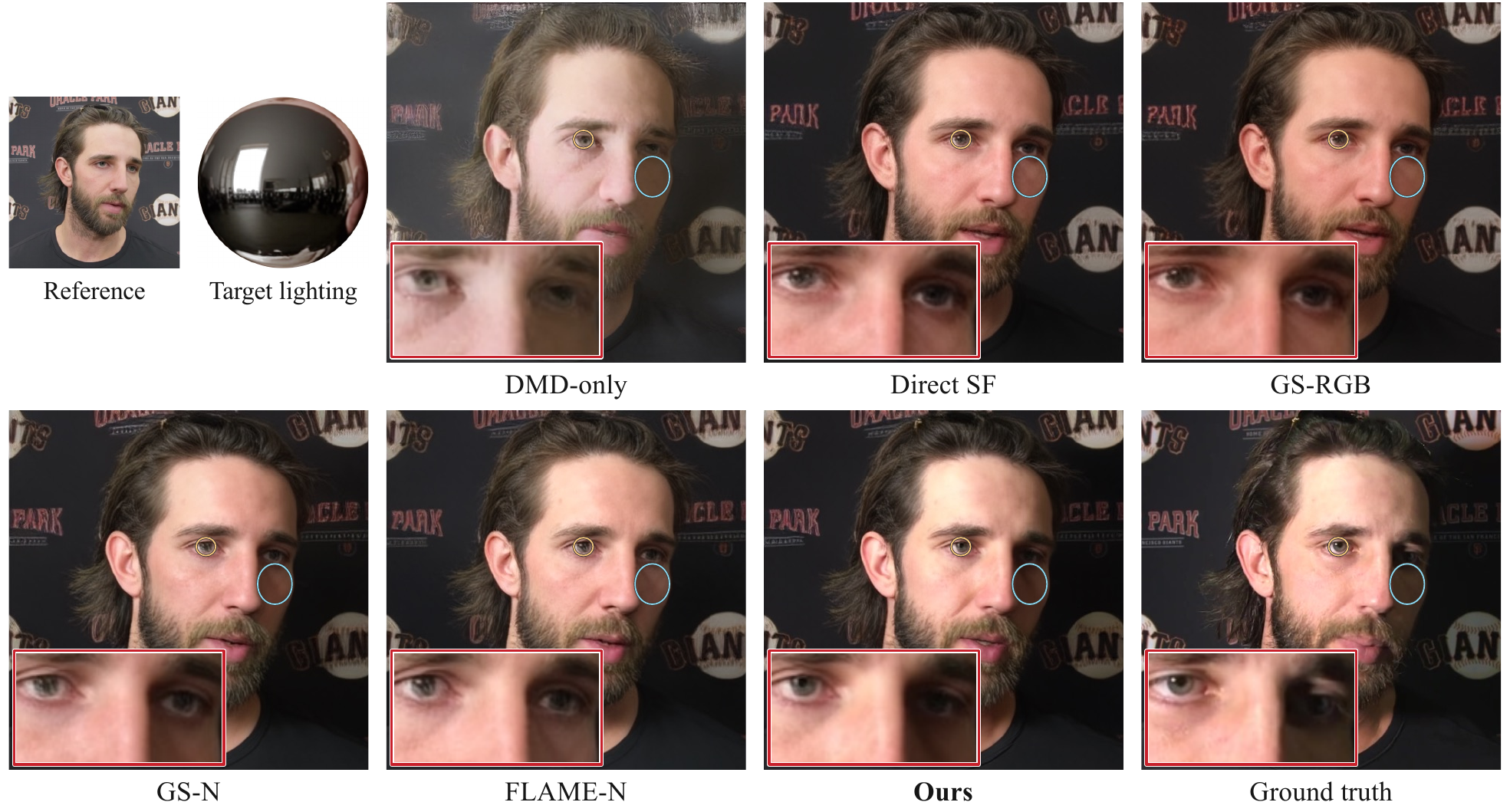}
    \caption{\textbf{Effect of training strategy and intrinsic proxy on VFHQ.}
Magnified insets highlight eye appearance and shadows around the nose.}
    \label{fig:supp_proxy_ablation}
\end{figure}

\vspace{-2mm}
\subsection{Heterogeneous Distillation}
\label{sec:supp_distillation}
Table~\ref{tab:supp_shader_interfaces} summarizes the teacher--student interface differences.
The student uses TAEHV for condition encoding, while the Relit-LiVE teacher~\cite{relitlive} and auxiliary score model retain Wan-VAE.
Since TAEHV is compatible with the Wan video latent space, student-generated video latents can be evaluated directly by both score models.

\noindent\textbf{ODE warm-up.}
From precomputed multi-step teacher ODE trajectories, we independently sample a trajectory state for each 3-frame latent chunk.
Given a noisy state from the teacher's ODE trajectory, the student predicts the trajectory's clean terminal video latent, conditioned on proxy-rendered PBR buffers, the reference portrait, and target illumination.
We minimize latent MSE between the prediction and this target.

\noindent\textbf{Distribution matching.}
Following DMD~\cite{yin2024one} and Self-Forcing~\cite{huang2026self}, clean video latents $\hat{\mathbf{z}}$ from self-forced student rollouts are perturbed into noisy latents $\mathbf{z}_{\sigma}$ at a sampled noise level $\sigma$:
\begin{equation}
    \mathbf{z}_{\sigma}
    = (1-\sigma)\hat{\mathbf{z}}+\sigma\boldsymbol{\epsilon},
    \qquad \boldsymbol{\epsilon}\sim\mathcal{N}(\mathbf{0},\mathbf{I}).
    \label{eq:supp_noising}
\end{equation}
Here, $\boldsymbol{\epsilon}$ denotes standard Gaussian noise and $\mathbf{I}$ is the identity covariance matrix.
To match the teacher's joint video--environment input, we encode the target log-HDR environment maps with Wan-VAE and perturb the resulting latent at the same noise level $\sigma$.
Both the frozen teacher and the trainable auxiliary score model receive the noisy video and environment latents, together with the full set of intrinsic conditions extracted by DiffusionRenderer~\cite{liang2025diffusionrenderer}.
We convert their video flow predictions into clean-latent estimates $\hat{\mathbf{z}}_{T}$ (teacher) and $\hat{\mathbf{z}}_{\phi}$ (auxiliary model with parameters $\phi$), and update the student via the surrogate loss:
\begin{equation}
\begin{aligned}
    \mathbf{g}
    &= \frac{\hat{\mathbf{z}}_{\phi}-\hat{\mathbf{z}}_{T}}
    {\tfrac{1}{M}\|\hat{\mathbf{z}}-\hat{\mathbf{z}}_{T}\|_{1}},\\
    \mathcal{L}_{\mathrm{DMD}}
    &= \mathbb{E}\!\left[\frac{1}{2|\mathcal{I}|}\Bigl\|
    \hat{\mathbf{z}}_{\mathcal{I}}-
    \operatorname{sg}[\hat{\mathbf{z}}_{\mathcal{I}}-\mathbf{g}_{\mathcal{I}}]
    \Bigr\|_{2}^{2}\right].
\end{aligned}
\label{eq:supp_dmd}
\end{equation}
Here, $\mathbf{g}$ is the normalized DMD gradient signal and $M$ is the number of scalar entries per video latent sample.
The set $\mathcal{I}$ indexes gradient-bearing rollout entries, $|\mathcal{I}|$ is their count, subscripts $\mathcal{I}$ select these entries, and $\operatorname{sg}[\cdot]$ denotes the stop-gradient operator.
The expectation averages over conditioning samples, student rollouts, noise levels, and Gaussian noise.
The auxiliary model is updated via flow matching on detached noisy student rollouts.

\subsection{Paired Portrait Adaptation}
\label{sec:supp_paired}
\noindent\textbf{Patch adversarial supervision.}
For $\mathcal{L}_{\mathrm{adv}}$, we apply a spectral-normalized 2D PatchGAN discriminator $D$~\cite{isola2017image} to all decoded RGB frames using the hinge objective:
\begin{equation}
\begin{aligned}
    \mathcal{L}_{D}
    &= \mathbb{E}[\max(0,1-D(Y))]\\
    &\quad+\mathbb{E}[\max(0,1+D(\hat{Y}))],\\
    \mathcal{L}_{\mathrm{adv}}
    &= -\mathbb{E}[D(\hat{Y})],
\end{aligned}
\label{eq:supp_patchgan}
\end{equation}
where $\mathcal{L}_{D}$ is the discriminator loss, $Y$ and $\hat{Y}$ are real and generated frames, and expectations average over video frames and spatial patches.

\newcommand{\SuppDirectRelightingRows}{%
\midrule
3DPR & 15.27 & 0.579 & 0.392 & 0.351 & 0.150 & 0.130 & 18.64 & 0.651 & 0.319 & 0.445 & 0.162 & \textbf{0.100} & 0.073 \\
Lite2Relight & 16.52 & 0.601 & 0.312 & 0.465 & 0.108 & 0.094 & 15.27 & 0.578 & 0.440 & 0.430 & 0.127 & 0.108 & 3.367 \\
DiffusionRenderer$^{\dagger}$ & 15.89 & 0.600 & 0.235 & 0.736 & 0.114 & 0.086 & 16.24 & 0.646 & 0.322 & 0.693 & 0.141 & 0.117 & 0.105 \\
Relit-LiVE$^{\dagger}$ & 17.14 & 0.661 & 0.237 & 0.629 & 0.109 & 0.080 & 14.83 & 0.652 & 0.304 & 0.683 & 0.125 & 0.104 & 0.056 \\
UniRelight & 18.00 & \textbf{0.730} & 0.188 & 0.724 & 0.090 & 0.066 & 14.12 & 0.639 & 0.318 & 0.637 & 0.135 & 0.107 & 0.136 \\
}

\begin{table*}[t]
\centering
\caption{\textbf{Quantitative comparison with direct relighting methods on VFHQ and FaceOLAT.}
Direct relighting methods take ground-truth target-pose RGB frames as input, whereas LuxAvatar jointly animates and relights a single reference portrait.
Best results are in bold.
FPS measures end-to-end throughput on a 57-frame clip on a single RTX 5090 D GPU.
$^{\dagger}$Uses DiffusionRenderer inverse rendering to extract intrinsic inputs.
$^{*}$With single-GPU inference optimization.}
\label{tab:supp_direct_relighting}
\begingroup
\small
\setlength{\tabcolsep}{1.25pt}
\renewcommand{\arraystretch}{1.1}
\begin{tabular*}{\textwidth}{@{\extracolsep{\fill}}l*{13}{c}@{}}
\toprule
& \multicolumn{6}{c}{VFHQ} & \multicolumn{6}{c}{FaceOLAT} & \\
\cmidrule(lr){2-7}\cmidrule(lr){8-13}\cmidrule(l){14-14}
Method & PSNR $\uparrow$ & SSIM $\uparrow$ & LPIPS $\downarrow$ & CSIM $\uparrow$ & AED $\downarrow$ & APD $\downarrow$ & PSNR $\uparrow$ & SSIM $\uparrow$ & LPIPS $\downarrow$ & CSIM $\uparrow$ & AED $\downarrow$ & APD $\downarrow$ & FPS $\uparrow$ \\
\SuppDirectRelightingRows
\midrule
\textbf{Ours} & \textbf{21.32} & 0.710 & \textbf{0.168} & \textbf{0.739} & \textbf{0.071} & \textbf{0.058} & \textbf{27.83} & \textbf{0.826} & \textbf{0.132} & \textbf{0.747} & \textbf{0.098} & 0.101 & \textbf{12.46 / 20.31}$^{*}$ \\
\bottomrule
\end{tabular*}
\par\endgroup
\end{table*}

\vspace{-2mm}
\subsection{Training Data and Model Selection}
\label{sec:supp_training_data}
\noindent\textbf{Data composition.}
For paired portrait adaptation, we use 1,000 synthesized relit video records, 40,355 reverse-constructed environment-map relighting pairs, 41,777 reverse-constructed background harmonization pairs, and 55,840 OLAT relit sequences.
These pools are sampled with weights of $5\%$, $45\%$, $20\%$, and $30\%$, respectively.
Synthesized relit and OLAT clips contain 57 frames, while reverse-constructed clips contain 81 frames.

\noindent\textbf{Data use across training stages.}
The intrinsic Gaussian proxy is trained on 46,795 VFHQ clip groups using original RGB frames and DiffusionRenderer intrinsic pseudo-labels.
ODE warm-up and DMD use VFHQ clips with sampled HDR environment maps.
The student receives proxy-rendered PBR buffers, while the teacher and auxiliary score model use DiffusionRenderer intrinsics.
ODE warm-up learns from teacher-generated trajectories, whereas DMD transfers the teacher's relighting prior without paired relit RGB targets.
For paired adaptation, the student continues to use proxy-rendered PBR buffers and is supervised by synthesized relit videos, original videos from reverse-constructed pairs, and OLAT relit sequences.

\noindent\textbf{Teacher selection.}
Relit-LiVE generally produces softer, more natural facial lighting.
DiffusionRenderer is more prone to producing an overly specular, metallic appearance on faces.
We therefore use Relit-LiVE as the distillation teacher to transfer its broad relighting prior to our shader.
However, Relit-LiVE may miss subtle expression changes, so we use DiffusionRenderer to synthesize paired relit targets that better preserve the source video's facial motion.
We visually inspect the synthesized relit videos before including them in the training set.

\noindent\textbf{Reference editing for reverse pairs.}
The edited reference should replace the original illumination, including its cast shadows, while the original video remains the real target.
In our experience, UniRelight~\cite{he2026unirelight} achieves more pronounced and realistic lighting changes than Qwen-Image-Edit~\cite{wu2025qwenimagetechnicalreport}, reducing residual source shading that would weaken relighting supervision.
We therefore use UniRelight for environment-map reference editing.
For background pairs, we favor Qwen-Image-Edit for its clean background extraction and coherent joint editing of backgrounds and portrait lighting.
Since these edits can alter identity, we use ArcFace~\cite{deng2019arcface} cosine similarity to flag potential identity changes, followed by visual review to exclude failed edits.

\setcounter{topnumber}{1}
\begin{table}[t]
\centering
\caption{\textbf{Quantitative background harmonization comparison on VFHQ.}
Best results are in bold.}
\label{tab:supp_background_harmonization}
\begingroup
\footnotesize
\setlength{\tabcolsep}{1.5pt}
\renewcommand{\arraystretch}{1.05}
\setbox0=\hbox{%
\begin{tabular*}{\linewidth}{@{\extracolsep{\fill}}l*{6}{c}@{}}
\toprule
Method & PSNR $\uparrow$ & SSIM $\uparrow$ & LPIPS $\downarrow$ & CSIM $\uparrow$ & AED $\downarrow$ & APD $\downarrow$ \\
\midrule
FYE + RelightVid & 11.59 & 0.377 & 0.547 & 0.510 & 0.137 & 0.122 \\
FYE + UniLumos & 11.98 & 0.374 & 0.507 & 0.386 & 0.141 & 0.116 \\
UIKA + RelightVid & 12.73 & 0.492 & 0.428 & 0.487 & 0.115 & 0.099 \\
UIKA + UniLumos & 13.07 & 0.450 & 0.399 & 0.387 & 0.113 & 0.094 \\
\midrule
\textbf{Ours} & \textbf{16.91} & \textbf{0.616} & \textbf{0.265} & \textbf{0.545} & \textbf{0.093} & \textbf{0.077} \\
\bottomrule
\end{tabular*}}
\ifdim\wd0>\linewidth
  \PackageError{bg-harmonization-table}{Table exceeds column width}{Reduce column spacing.}
\fi
\box0
\par\endgroup
\end{table}

\vspace{-2mm}
\section{Streaming and Inference Optimization}
\label{sec:supp_streaming}
\label{sec:supp_optimization}

\noindent\textbf{Bounded attention context.}
\label{sec:supp_context}
To keep attention cost bounded during streaming, each chunk attends to itself, its immediate predecessor, and up to two uniformly spaced chunks from the earlier history.
The reference and optional background prefixes remain accessible throughout generation and do not count toward this four-chunk budget.
Numbering chunks from 1, we select $m=\min(2,k-2)$ earlier chunks for the current chunk $k>2$, with indices:
\begin{equation}
    h_j = 1+\left\lfloor
    \frac{(j+\tfrac{1}{2})(k-2)}{m}
    \right\rfloor,
    \quad j=0,\ldots,m-1,
    \label{eq:supp_history_selection}
\end{equation}
where $h_j$ is the index of the $j$-th selected history chunk and $\lfloor\cdot\rfloor$ denotes the floor operation.
For the first two chunks, all available video chunks are used.
Selected chunks retain their original temporal indices under RoPE.

After the two denoising steps, we run a clean-latent forward pass to cache the completed chunk's keys and values for subsequent chunks, following Self-Forcing~\cite{huang2026self}.
This extra pass can be skipped for the final chunk of the sequence.

\noindent\textbf{Condition caching and kernel fusion.}
We compute reference, optional background, and empty-prompt text projections once per stream and cache them across chunks.
For each chunk, we compute lighting projections and joint text--lighting cross-attention K/V once, sharing them across the two denoising steps and the clean-latent KV update.
We also fuse QKV projections, RoPE, AdaLN modulation, and gated residual updates using \texttt{torch.compile} and custom CUDA kernels.


\noindent\textbf{Exact tiered KV management.}
Historical chunks may be selected again by Eq.~\eqref{eq:supp_history_selection}, so their KV states must remain available until their last use.
Since this selection depends only on chunk indices, we can determine cache lifetimes without accessing future frames.
For short clips of known length, we precompute the access schedule and recycle GPU slots after each chunk's last use.
For long streams, inactive KV blocks are offloaded to CPU memory and reloaded on demand, keeping GPU memory bounded.
Under our selection rule, chunk $i$ can be discarded once generation advances beyond chunk $4i+1$, independently of the final sequence length.
These storage operations preserve the BF16 cache values and the selected attention context.

\section{Additional Evaluation}
\label{sec:supp_evaluation}

\subsection{Environment-Map Relighting}
\label{sec:supp_env_qualitative}
Figure~\ref{fig:supp_env_relighting} presents additional qualitative comparisons with the joint animation-and-relighting baselines from the main paper.
Across varied identities and target illumination, our method better preserves facial appearance and produces more plausible directional shading while following the driving expressions.
\begin{table}[!htbp]
\centering
\caption{\textbf{User study mean ranks over 21 questionnaires.}
Lower is better, best results are in bold.}
\label{tab:supp_user_study}
\begingroup
\footnotesize
\setlength{\tabcolsep}{2pt}
\renewcommand{\arraystretch}{1.05}
\setbox0=\hbox{%
\begin{tabular*}{\linewidth}{@{\extracolsep{\fill}}lccc@{}}
\toprule
\multicolumn{4}{c}{Environment Map Relighting} \\
\midrule
Method & Quality $\downarrow$ & Lighting $\downarrow$ & Identity $\downarrow$ \\
\midrule
CapHuman & 4.90 & 4.83 & 4.63 \\
ControlFace & 4.33 & 4.40 & 4.07 \\
LCVD & 4.37 & 4.40 & 4.27 \\
UIKA + UniRelight & 2.27 & 2.43 & 2.47 \\
UIKA + Relit-LiVE & 4.03 & 3.77 & 4.37 \\
\textbf{Ours} & \textbf{1.10} & \textbf{1.17} & \textbf{1.20} \\
\midrule
\multicolumn{4}{c}{Background Harmonization} \\
\midrule
Method & Quality $\downarrow$ & Background $\downarrow$ & Identity $\downarrow$ \\
\midrule
FYE + RelightVid & 4.03 & 4.13 & 4.33 \\
FYE + UniLumos & 3.53 & 3.13 & 3.20 \\
UIKA + RelightVid & 3.90 & 3.90 & 3.80 \\
UIKA + UniLumos & 2.30 & 2.60 & 2.40 \\
\textbf{Ours} & \textbf{1.23} & \textbf{1.23} & \textbf{1.27} \\
\bottomrule
\end{tabular*}}
\ifdim\wd0>\linewidth
  \PackageError{user-study-table}{Table exceeds column width}{Reduce column spacing.}
\fi
\box0
\par\endgroup
\end{table}

\subsection{Direct Environment-Map Relighting}
\label{sec:supp_direct_relighting}
Table~\ref{tab:supp_direct_relighting} compares LuxAvatar with five direct relighting baselines: 3DPR~\cite{rao20253dpr}, Lite2Relight~\cite{rao2024lite2relight}, DiffusionRenderer~\cite{liang2025diffusionrenderer}, Relit-LiVE~\cite{relitlive}, and UniRelight~\cite{he2026unirelight}.
These baselines receive target-pose RGB frames as input.
For DiffusionRenderer and Relit-LiVE, we first use DiffusionRenderer's inverse rendering model to extract intrinsic buffers from these frames for subsequent relighting.
LuxAvatar instead synthesizes both target head motion and novel illumination from a single reference portrait.
Figure~\ref{fig:supp_direct_relighting} provides qualitative comparisons.

Despite the baselines' access to target-pose RGB frames, LuxAvatar performs best in PSNR, LPIPS, CSIM, and AED on both benchmarks.
With inference optimization, LuxAvatar achieves approximately $6$--$360\times$ the FPS of these baselines.
In particular, paired portrait adaptation substantially improves facial shading fidelity and color accuracy over general-scene video relighting methods, especially on physically captured FaceOLAT sequences.

\subsection{Background Harmonization}
\label{sec:supp_background}
In our harmonization comparison against two-stage cascades (Follow-Your-Emoji (FYE)~\cite{ma2026follow} or UIKA~\cite{wu2026uika} followed by RelightVid~\cite{fang2025relightvid} or UniLumos~\cite{liu2026unilumos}), the driven video from each animation method is shared across both downstream harmonization methods under identical source, motion, and background inputs.
We evaluate on VFHQ test clips, following the evaluation protocol detailed in the main paper.
As shown in Table~\ref{tab:supp_background_harmonization}, LuxAvatar outperforms all four cascades across reconstruction, identity, and motion metrics.
Figure~\ref{fig:supp_background_harmonization} provides additional qualitative comparisons across diverse target backgrounds.

\begin{figure}[t]
\centering
    \includegraphics[width=\linewidth]{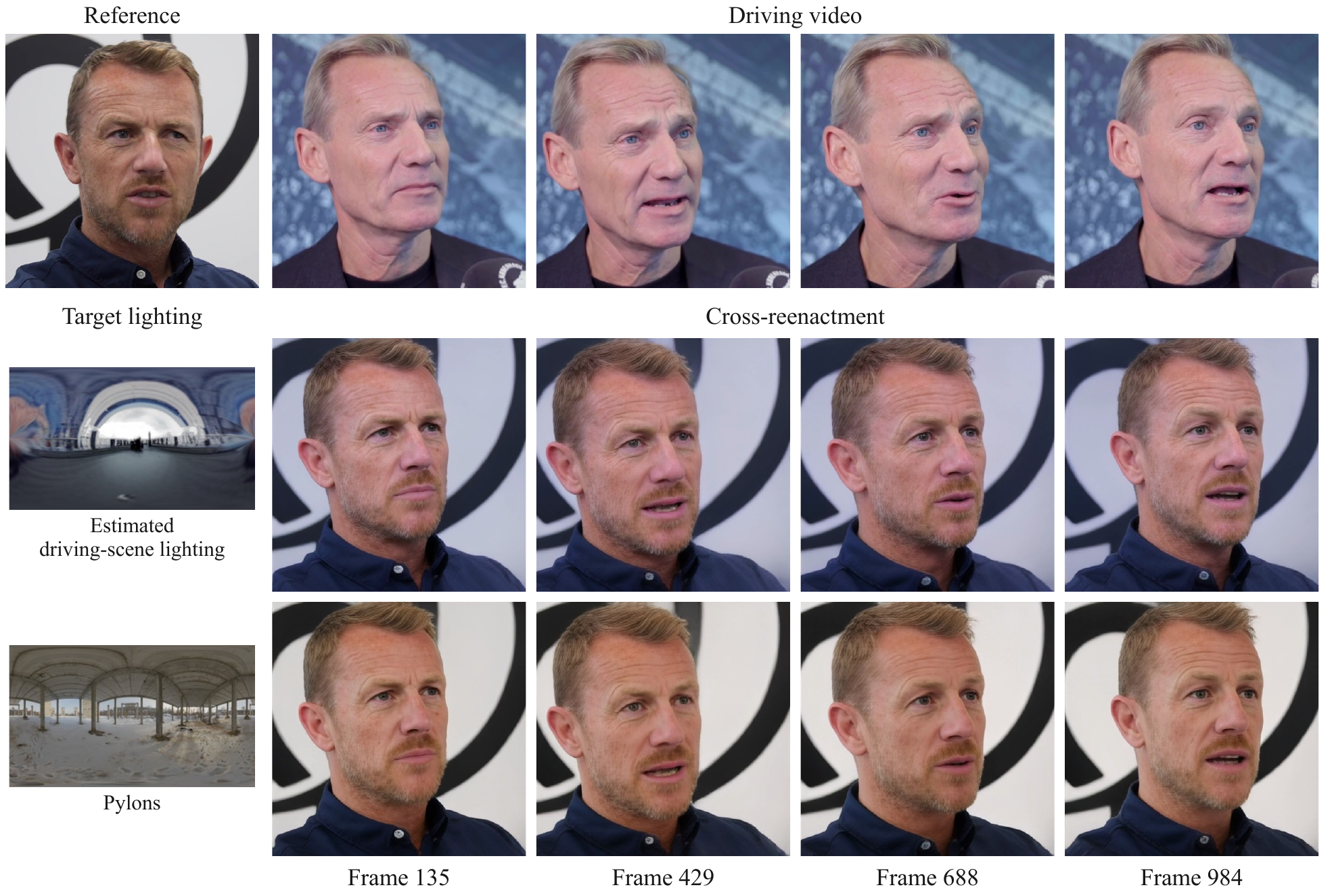}
    \caption{\textbf{Long-sequence cross-reenactment under two lighting conditions.}
The top row shows the driving frames, while the middle and bottom rows show our results under the estimated driving-scene illumination and the Pylons environment map.}
    \label{fig:supp_long_sequence}
\end{figure}

\vspace{-2mm}
\subsection{Training Strategy and Intrinsic Proxy}
\label{sec:supp_proxy_ablation}
Figure~\ref{fig:supp_proxy_ablation} complements the quantitative ablations in the main paper.
DMD-only loses fine facial details, while Direct SF and the normal-only variants show discrepancies in eye appearance and shadows around the nose.
Our full model better preserves these details under target lighting.

\subsection{Long-Sequence Stability}
\label{sec:supp_long_sequence}
Figure~\ref{fig:supp_long_sequence} shows cross-reenactment over continuous 984-frame rollouts.
We animate the same reference identity using another subject's FLAME motion and camera parameters under two fixed environment maps, generating each sequence without resetting the KV cache.
The displayed frames retain consistent identity and distinct lighting while following changes in the driving expression.

\subsection{User Study}
\label{sec:supp_user_study}
We evaluate 20 video cases: 10 for environment-map relighting with six methods and 10 for background harmonization with five methods.
The interface displays synchronized, looping videos alongside the reference portrait, driving video, and target illumination or background.
Case order and anonymous method labels are randomized for each session.
For each case, respondents rank all methods from best to worst, without ties, on overall visual quality, lighting or background consistency, and identity preservation.

Table~\ref{tab:supp_user_study} summarizes mean ranks over 21 completed questionnaires, with the two tasks evaluated separately.
LuxAvatar achieves the lowest mean rank for overall visual quality, lighting or background consistency, and identity preservation in both tasks.

\section{Limitations}
\label{sec:supp_limitations}
Two aspects of LuxAvatar remain to be addressed in future work.
The FLAME-driven proxy lacks an explicit tongue model, so tongue motions such as protrusion cannot be directly controlled.
In addition, our training data lack multi-view coverage of the entire head, leaving the geometry and appearance of the back of the head underconstrained and limiting reliable synthesis of rear views.


\end{document}